%% file: ICVGIP-Latex-Template.tex
\documentclass[sigconf]{acmart}
\usepackage{soul}
\usepackage{booktabs} 
\usepackage{array}
\usepackage{subcaption}

\setcopyright{none}
\setcopyright{rightsretained}

\copyrightyear{2022}
\acmYear{2022}
\setcopyright{acmcopyright}\acmConference[ICVGIP'22]{Proceedings of the Thirteenth Indian Conference on Computer Vision, Graphics and Image Processing}{December 8--10, 2022}{Gandhinagar, India}
\acmBooktitle{Proceedings of the Thirteenth Indian Conference on Computer Vision, Graphics and Image Processing (ICVGIP'22), December 8--10, 2022, Gandhinagar, India}
\acmPrice{15.00}
\acmDOI{10.1145/3571600.3571629}
\acmISBN{978-1-4503-9822-0/22/12}

\acmPrice{15.00}
\begin{document}
\title{A Dataset and Model for Crossing Indian Roads}

\author{Siddhi M Brahmbhatt}
\affiliation{%
  \institution{G H Patel College of Engineering and Technology}
  \city{Vallabh Vidyanagar, Anand}
  \state{Gujarat}
  \country{India}
  \postcode{388120}
}

\renewcommand{\shortauthors}{}

\begin{abstract}
Roads in medium-sized Indian towns often have lots of traffic but no (or disregarded) traffic stops. This makes it hard for the blind to cross roads safely, because vision is crucial to determine when crossing is safe. Automatic and reliable image-based safety classifiers thus have the potential to help the blind to cross Indian roads. Yet, we currently lack datasets collected on Indian roads from the pedestrian point-of-view, labelled with road crossing safety information. Existing classifiers from other countries are often intended for crossroads, and hence rely on the detection and presence of traffic lights, which is not applicable in Indian conditions. We introduce INDRA (INdian Dataset for RoAd crossing), the first dataset capturing videos of Indian roads from the pedestrian point-of-view. INDRA contains 104 videos comprising of 26k 1080p frames, each annotated with a binary road crossing safety label and vehicle bounding boxes. We train various classifiers to predict road crossing safety on this data, ranging from SVMs to convolutional neural networks (CNNs). The best performing model DilatedRoadCrossNet is a novel single-image architecture tailored for deployment on the Nvidia Jetson Nano. It achieves 79\% recall at 90\% precision on unseen images. Lastly, we present a wearable road crossing assistant running DilatedRoadCrossNet, which can help the blind cross Indian roads in real-time. The project webpage is \url{https://roadcross-assistant.github.io/Website/index.html}
\end{abstract}

%
%



\begin{teaserfigure}
\includegraphics[scale=0.5]{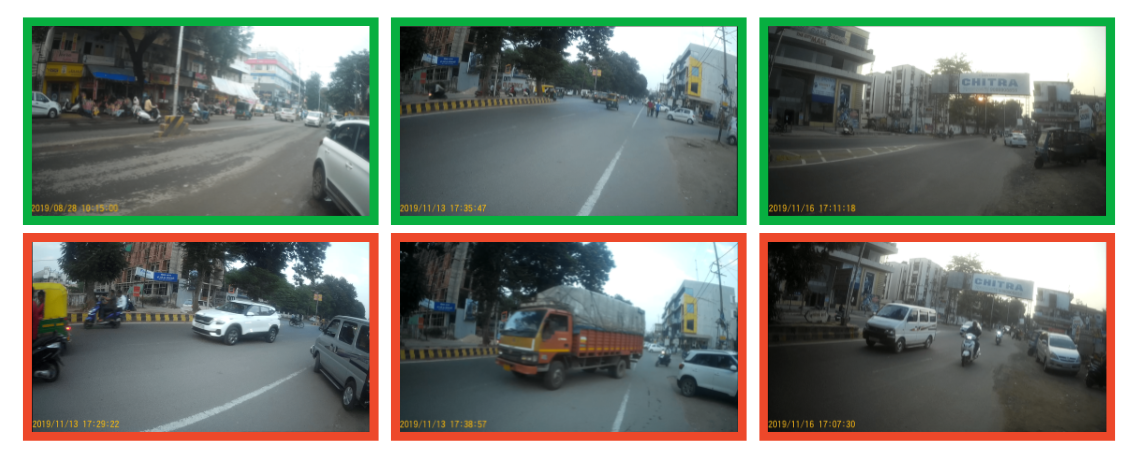}
\centering
\caption{Example frames from INDRA dataset (green=safe, red=unsafe). INDRA dataset consists of 104 videos (>26k frames) recording Indian traffic scenarios from pedestrian viewpoint. Each frame of the dataset is labelled as safe/unsafe for crossing road. Additionally, vehicle bounding boxes detected by state of the art RetinaNet for each frame of the dataset are released as auxiliary data.}
\Description{figure description}
\end{teaserfigure}

\maketitle
\pagestyle{empty}

\input{samplebody-conf}

\bibliographystyle{ACM-Reference-Format}
\bibliography{acmart}


\end{document}

%% file: samplebody-conf.tex
\begin{table*}
  \caption{Comparison with existing Indian road datasets. INDRA stands out for its pedestrian view-point, and binary road crossing safety labels}
  \begin{tabular}{c c c c c}
    \toprule
    Dataset & \#Images/ \#Sequences & Pedestrian view-point & Resolution  & Labels\\
    \midrule
    INDRA (ours) & 26K / 104 & \checkmark & 1920x1080 &  Binary road crossing safety \\ 
    IDD \cite{varma2019idd} & 10K / 180 &  & 1678x968  & Semantic segmentation \\
    DIRS21 \cite{esy0-sm56-21} & 5093 / - &  & 1920x1080 &  Object detection \\
  \bottomrule
\end{tabular}
\end{table*}

\section{Introduction}
Computer Vision is one of the advanced AI domains that allows computers to see and understand visual data (images/videos) in a way similar to how humans do. Advancements in computer vision research has revolutionized the usefulness of AI applications in daily lives. Also, with an increase in computational capacity along with research for developing neural networks with low latency, deep learning models can be deployed on mobile devices- making them accessible and affordable to the common people. One such field which has reaped significant benefits from these advancements is assistive technology for the visually impaired people.

A major problem faced by visually impaired people all over the world is crossing roads. White canes or walking sticks can help them scan their surroundings to detect any obstacles and/or orientation marks. But road crossing is a very sensitive activity, and requires a quick decision from pedestrian which is beyond the scope of normal walking sticks or white canes.

Companies like eSight \cite{eSight}, Seeing AI \cite{SeeingAI}, and the Sound of Vision System \cite{Caraiman_2017_ICCV} have developed some useful products for visually impaired people over the years; but they are not specifically designed for helping them cross roads. Also, these products are very costly so they are not feasible for use in developing countries. Low cost is a very essential factor to create a large scale impact in developing countries like India. Also, one major constraint while crossing roads in India (and other developing countries) is that there are no traffic lights or pedestrian crosswalks in small towns and small internal roads. So, our work in this paper is an attempt to develop a model to predict road crossing safety on Indian roads.

To summarize, we make the following contributions:
\begin{enumerate}
    \item Data: This paper introduces INDRA (INdian Dataset for RoAd crossing), a first of its kind dataset capturing Indian traffic scenarios (from pedestrian point-of-view) labelled with road crossing safety information.
    \item Prediction models: This paper presents analysis of rigorous training experiments performed to develop a classifier to predict road crossing safety.
    \item Working prototype: This paper also presents a low-cost and feasible working prototype of road crossing assistant which can potentially be used by visually impaired people to cross roads in real-time.
\end{enumerate}

\section{Related Work}

To the best of our knowledge, all the existing attempts to help visually impaired people cross roads are based on detection and localization of pedestrian crosswalk signs and traffic lights (hence applicable in developed countries only). In this section, we summarize some of these attempts.

One attempt to help visually impaired people cross roads has been the use of acoustic pedestrian lights \cite{harkey2007accessible}. These traffic lights use audible tones, verbal messages and/or vibrating surfaces to notify the pedestrians about useful information such as the color of traffic light and road crossing safety. Though this concept appears to be very useful and reliable, it is implemented only in urban areas. Also, there is almost no presence of acoustic lights in developing countries like India. 

Recent years have witnessed significant improvements in the task of detecting traffic lights for autonomous driving \cite{almeida2018traffic, li2021improved, gong2010recognition}. These advancements have been extended to improve the detection of pedestrian traffic lights for navigation support of visually impaired people \cite{martinez2017using}. But in developing countries like India, crossing roads is something much beyond just detecting pedestrian crosswalks and traffic lights. Initiatives like NavAssistAI \cite{NavAssistAI} and XIMIRA \cite{XIMIRA} are noteworthy when reviewing the existing solutions to help blind people cross roads. But again, they are not feasible for Indian traffic scenarios.

There are India road datasets available for autonomous navigation (see Table 1 for comparison), but they are not recorded from a pedestrian’s point of view. Also, to the best of our knowledge, no existing Indian road dataset is labelled for road crossing safety.

\section{Data}

Any dataset suitable for the task of predicting road crossing safety on Indian roads did not exist. There are  datasets available for semantic segmentation and object detection (see Table 1) for autonomous driving, but they are not recorded from a pedestrian’s point of view, and neither are they labelled for road crossing safety. So we introduce our own dataset: \textit{\textbf{INDRA (INdian Dataset for RoAd crossing)}}. Our dataset consists of 104 videos ($>$26K frames) capturing various traffic patterns from roads of a town of Gujarat, India. We collected the videos using head-mounted SJCam SJ4000 12MP action camera (Figure 2). Video specifications: MOV format, 1920*1080 resolution, 30 fps. 

\begin{figure}[htp]
    \centering
    \includegraphics[width=0.5\columnwidth]{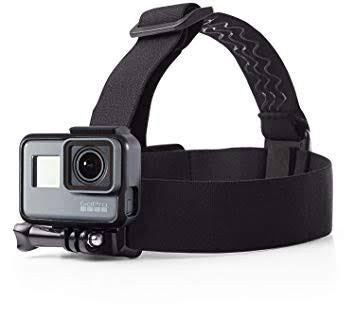}
    \caption{SJCam SJ4000 12MP action camera}
\end{figure}


We also manually annotated the videos with timestamps for durations safe for crossing (example images in Figure 1). For annotations, we had made a simple key-press based utility software: the annotator had to press pre-defined keys to store the safe-start and safe-end durations (in terms of frame number) in a csv file. The software also had provision to store multiple safe durations from a given video. We also release vehicle bounding boxes detected by state of the art RetinaNet for each frame of INDRA as auxiliary data for the community to use in further research.

\section{Problem Statement}

The task is to train models to predict road crossing safety using the dataset mentioned above. More precisely, given a sequence of k frames (say, $f_{n}$ to $f_{n-k}$), the goal is to predict whether $f_{n+1}$ is safe (here $f_{n}$ represents $n^{th}$ frame of the sequence). Precision-recall characteristics can be used to evaluate such classifiers \cite{hossin2015review, goutte2005probabilistic}. To elaborate:
\begin{enumerate}
    \item \textbf{Precision score} represents the proportion of positive predictions made by the model which are actually positive (i.e., high precision implies low number of false positives). It is extremely important to have high precision because the safety of the user depends on a low number of false positives.
    
    \item \textbf{Recall score} represents the proportion of positive predictions identified correctly by the model (i.e., high recall implies low number of false negatives). While high recall score is not the foremost priority in this application, higher recall is better because in the greed of getting a high precision score, we do not want the model to predict every frame as unsafe for crossing the road.
    
\end{enumerate}

\section{Machine Learning Methods}

In this section, we present the details of support vector machines (SVMs) trained on hand crafted features extracted from video frames of INDRA. We used 80 videos for training and remaining 24 videos for testing the trained SVMs. 
\subsection{Single frame SVM}
It is our simplest approach where we extracted simple per-frame features capturing number, location, size, speed and direction of vehicles. Further, we trained SVM classifier to predict whether a particular frame is safe for crossing road. For extracting features from a particular frame, we divided the video frame into 6 regions as shown in Figure 3.
\begin{figure}[htp]
    \centering
    \includegraphics[width=7.5cm]{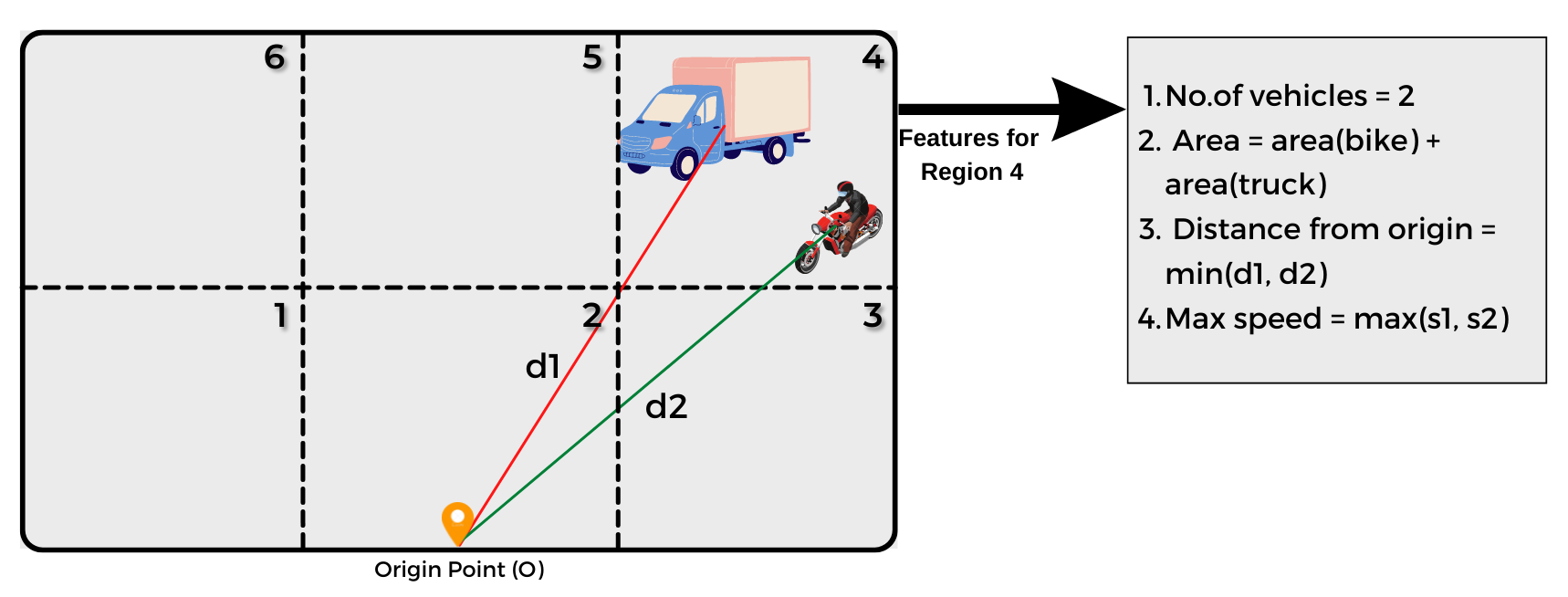}
    \caption{Feature extraction for single frame SVM model (Note: s1 = speed of truck \& s2 = speed of bike)}
\end{figure}

 It is very intuitive because even we as humans, when crossing roads, consciously/unconsciously consider information about vehicles in different regions of our field of vision. We extract following features from each of the 6 regions:

\begin{enumerate}
    \item Number of vehicles in the region.
    \item Total area covered by vehicles in the region
    \item Distance (from origin) of the vehicle closest to the origin. Here, the origin approximately represents the position of user with head mounted camera.
    \item Maximum speed of any vehicle in the region.
\end{enumerate}
Hence, a total of 6 * 4 = 24 features are extracted from each frame.
\begin{figure}[htp]
    \centering
    \includegraphics[width=\columnwidth]{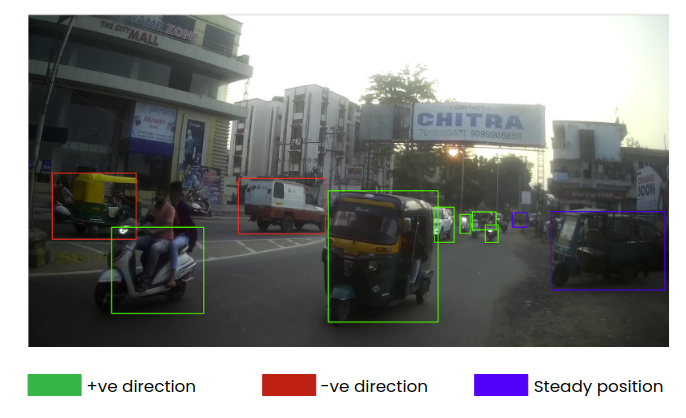}
    \caption{Directional vehicle filtering using KCF tracker}
\end{figure}

\begin{figure}[htp]
    \centering
    \includegraphics[width=\columnwidth]{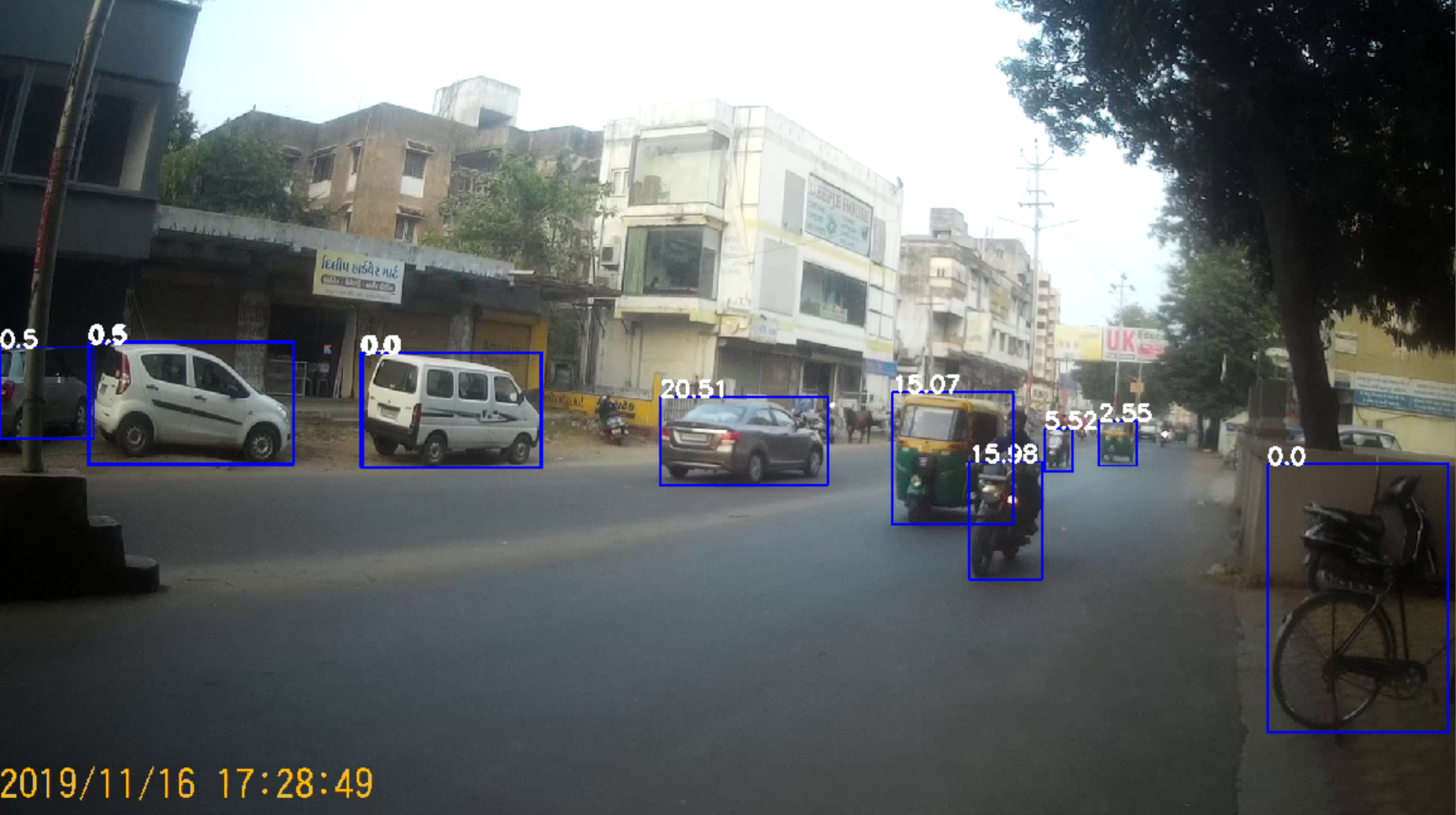}
    \caption{Vehicle speed detection using KCF tracker}
\end{figure}

We also applied directional vehicle filtering on these features. As seen in Figure 4, for crossing road, the information of vehicles on the other side of the divider (i.e., the vehicles marked in red) is not required. So, we did not consider those vehicles during feature extraction. 
To calculate the number, area and distance of vehicles, we used ImageAI library for video object detection using RetinaNet \cite{lin2017focal} to generate and save the vehicle bounding boxes. To get direction and speed (see Figure 5) of these vehicle bounding boxes, we used OpenCV's KCF (Kernelized Correlation Filters) \cite{henriques2014high} tracker. We applied minmax scaling to the extracted features and trained an SVM on the train set.

\subsection{Multi frame SVM}
As it is obvious that even we as humans do not decide whether it is safe to cross a road by just having one glance at the road, we extended the above approach and extracted multi-frame features instead of per-frame features.
\begin{figure}[htp]
    \centering
    \includegraphics[width=\columnwidth]{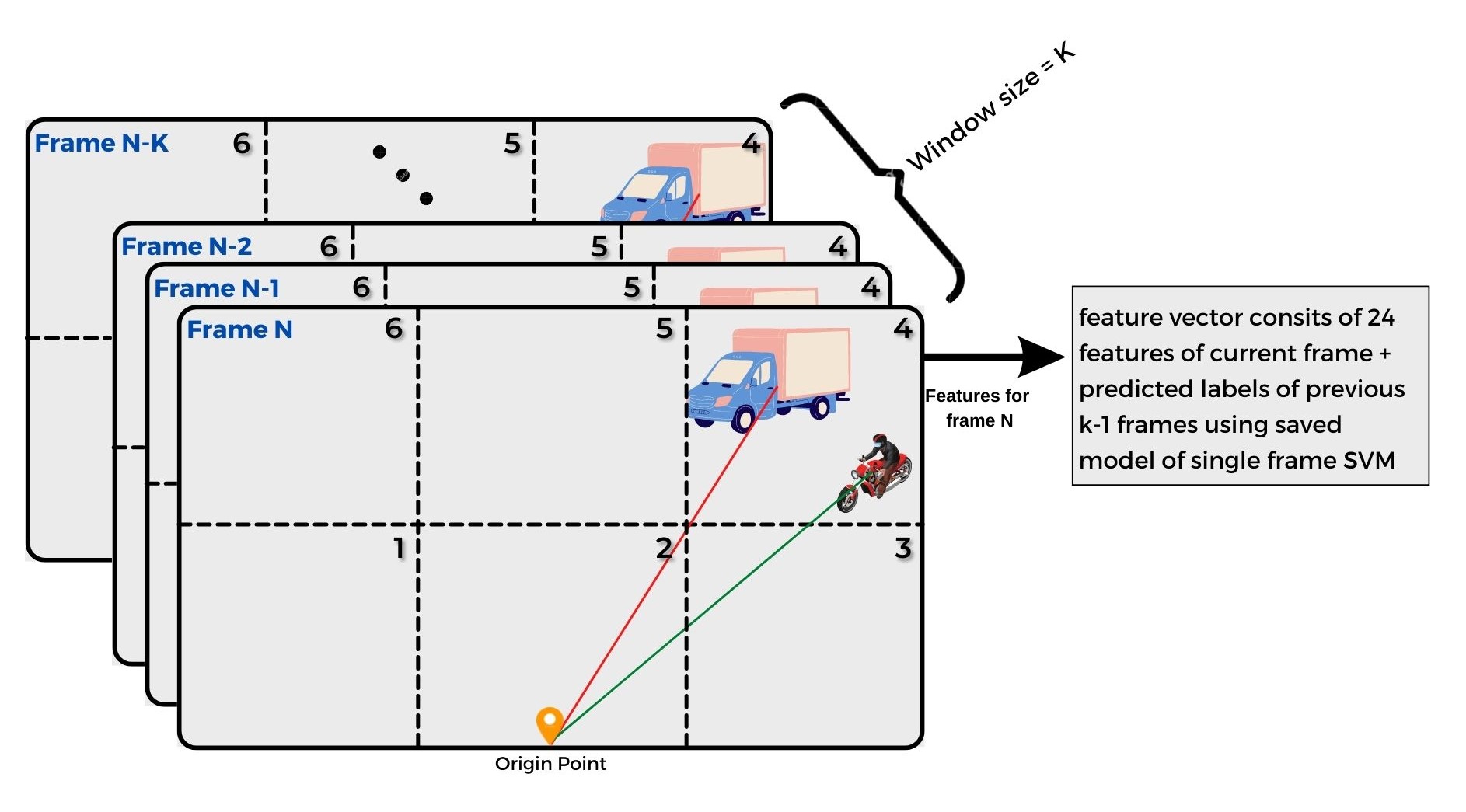}
    \caption{Feature extraction for multi frame SVM model}
\end{figure}

\begin{figure*}
    \centering
    \subfloat{\includegraphics[width=0.15\textwidth]{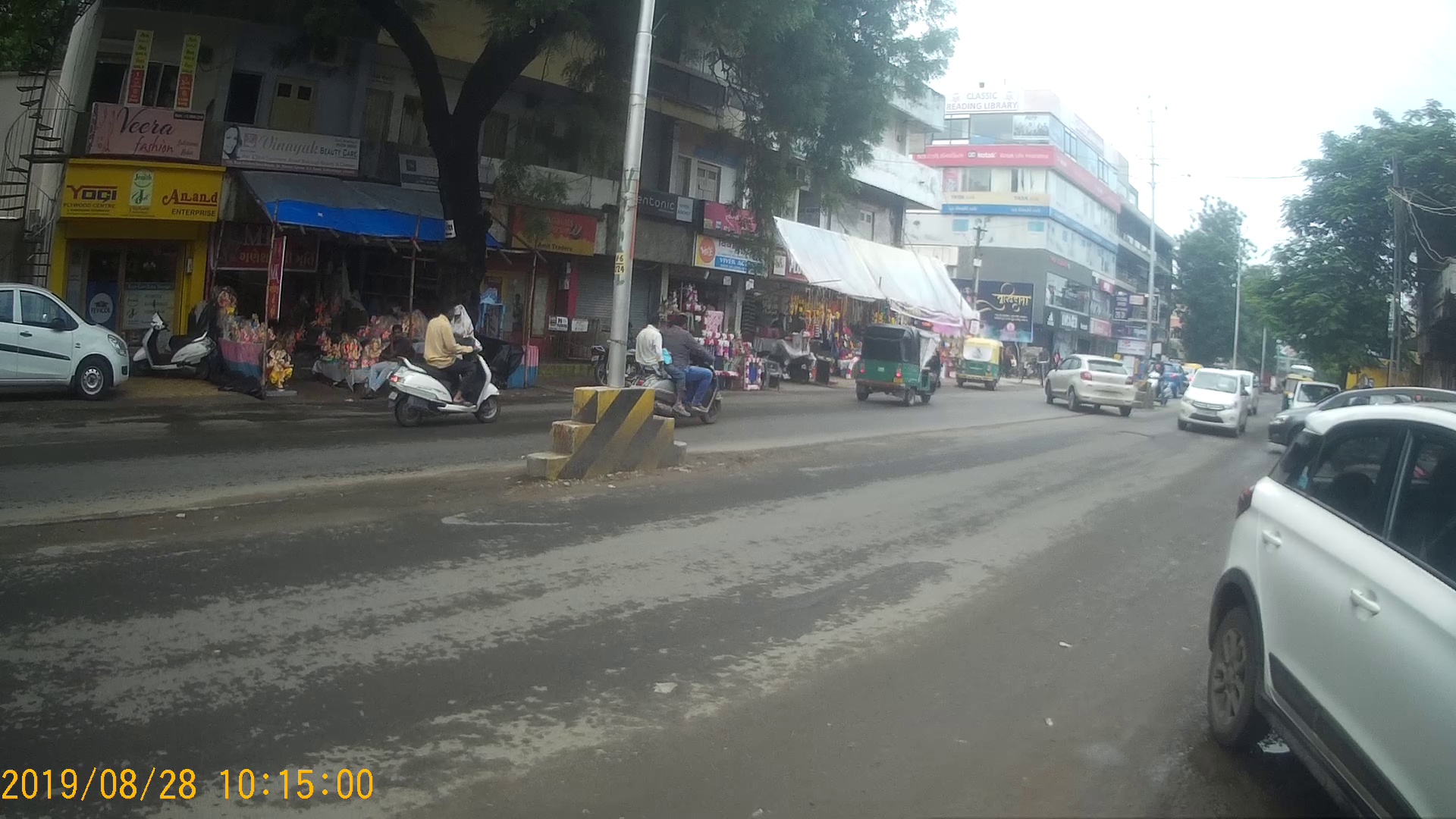}}\hfil
    \subfloat{\includegraphics[width=0.15\textwidth]{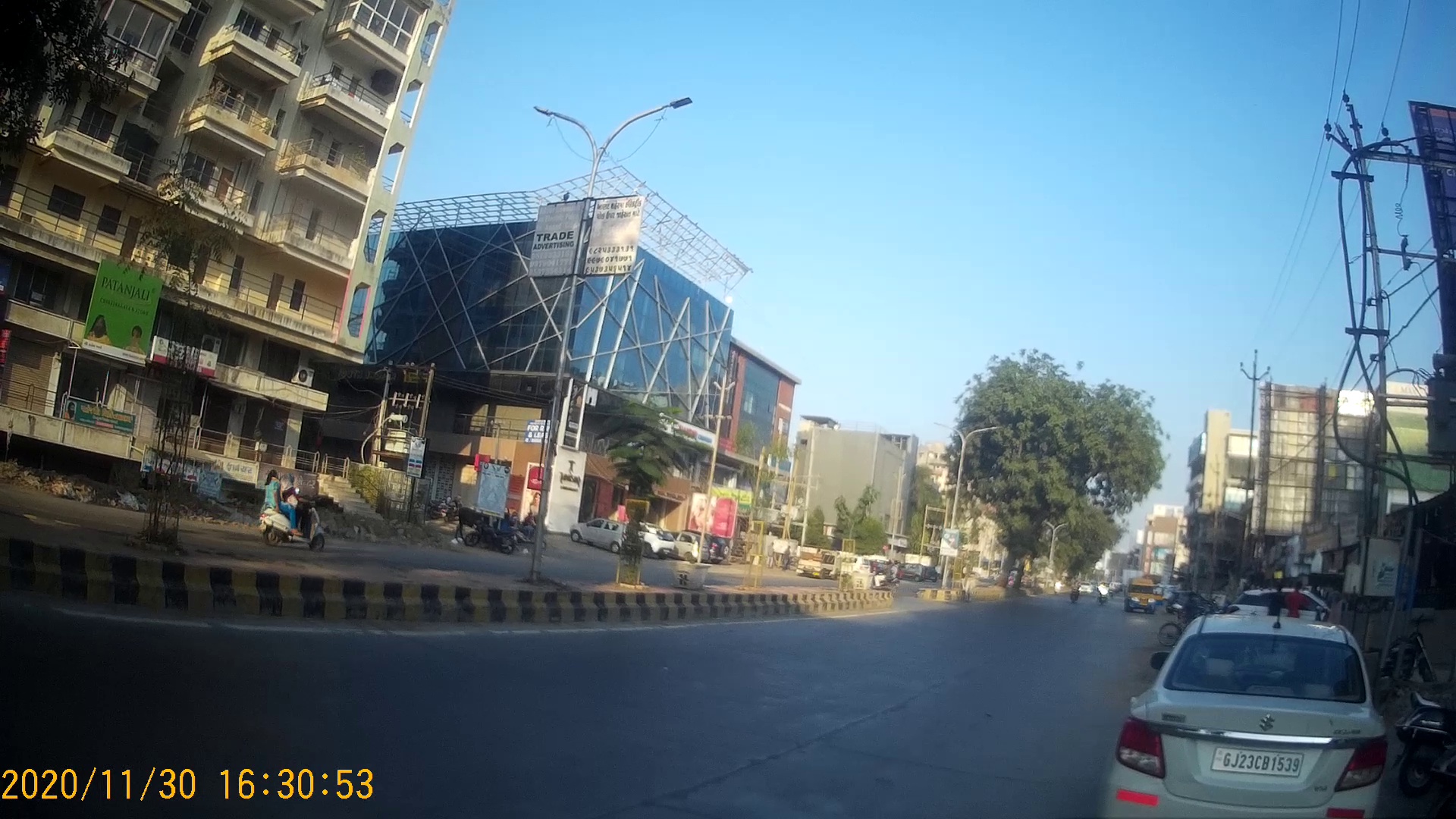}} \hfil
    \subfloat{\includegraphics[width=0.15\textwidth]{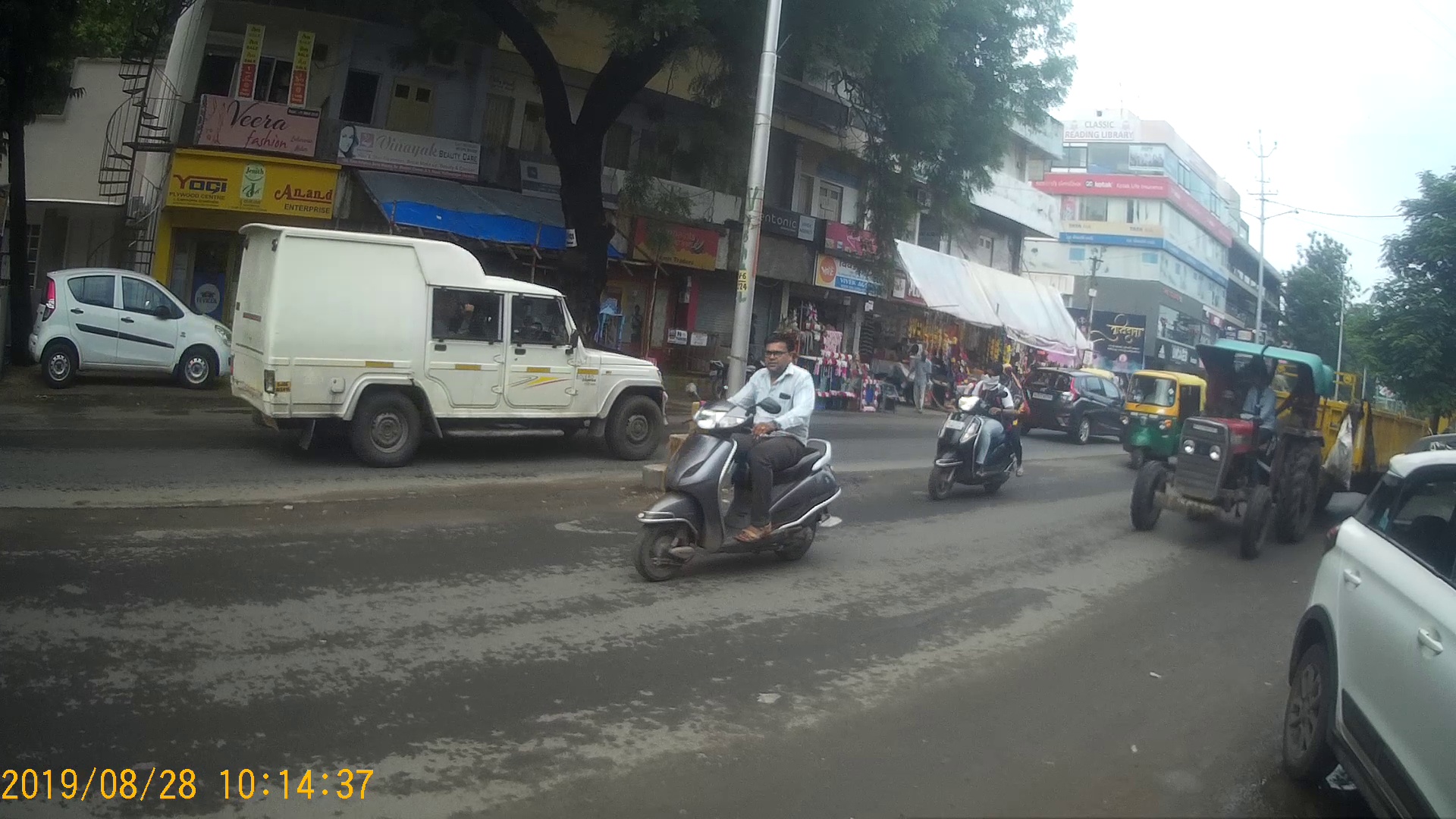}}\hfil
    \subfloat{\includegraphics[width=0.15\textwidth]{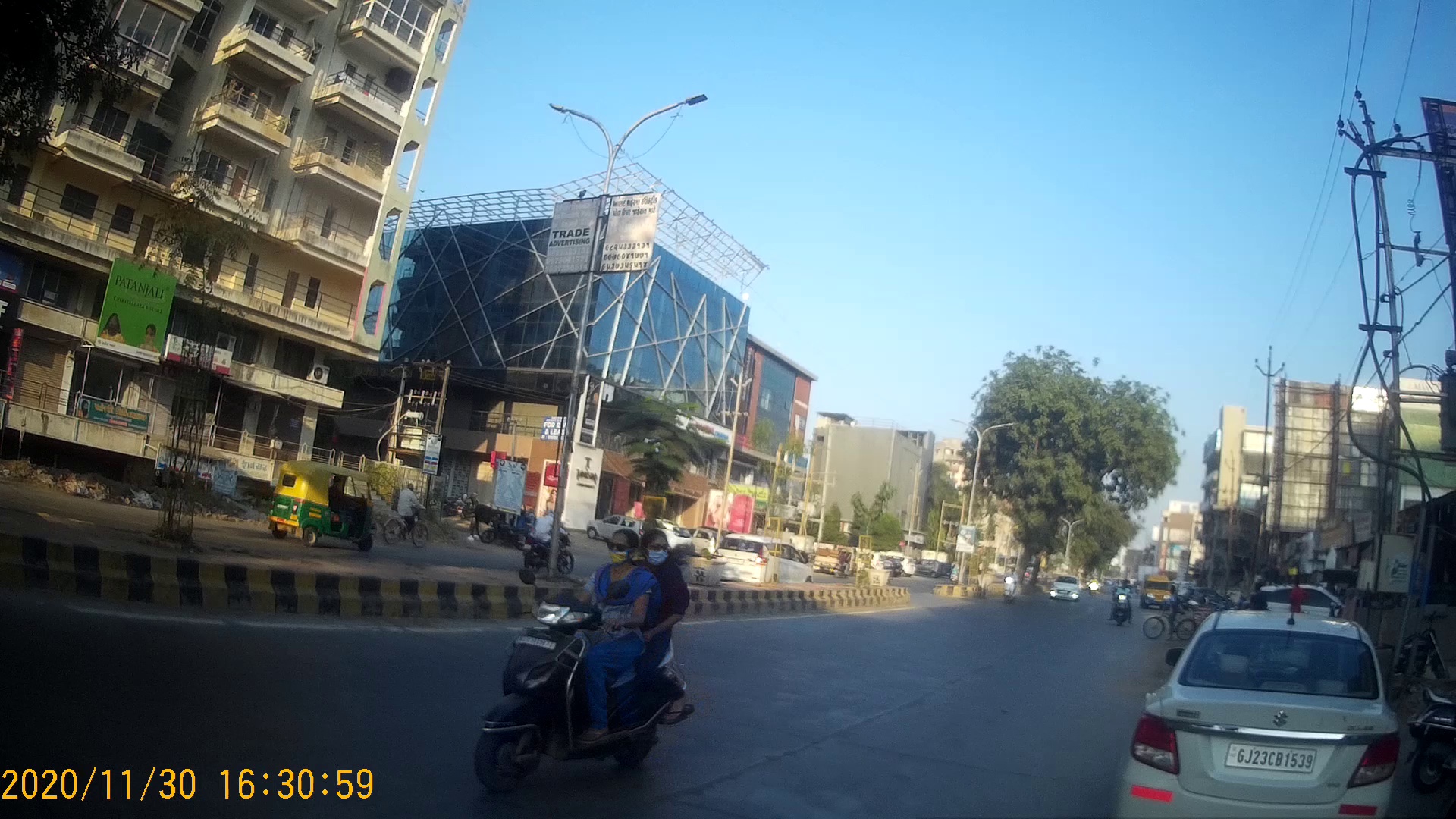}}\hfil
    \subfloat{\includegraphics[width=0.15\textwidth]{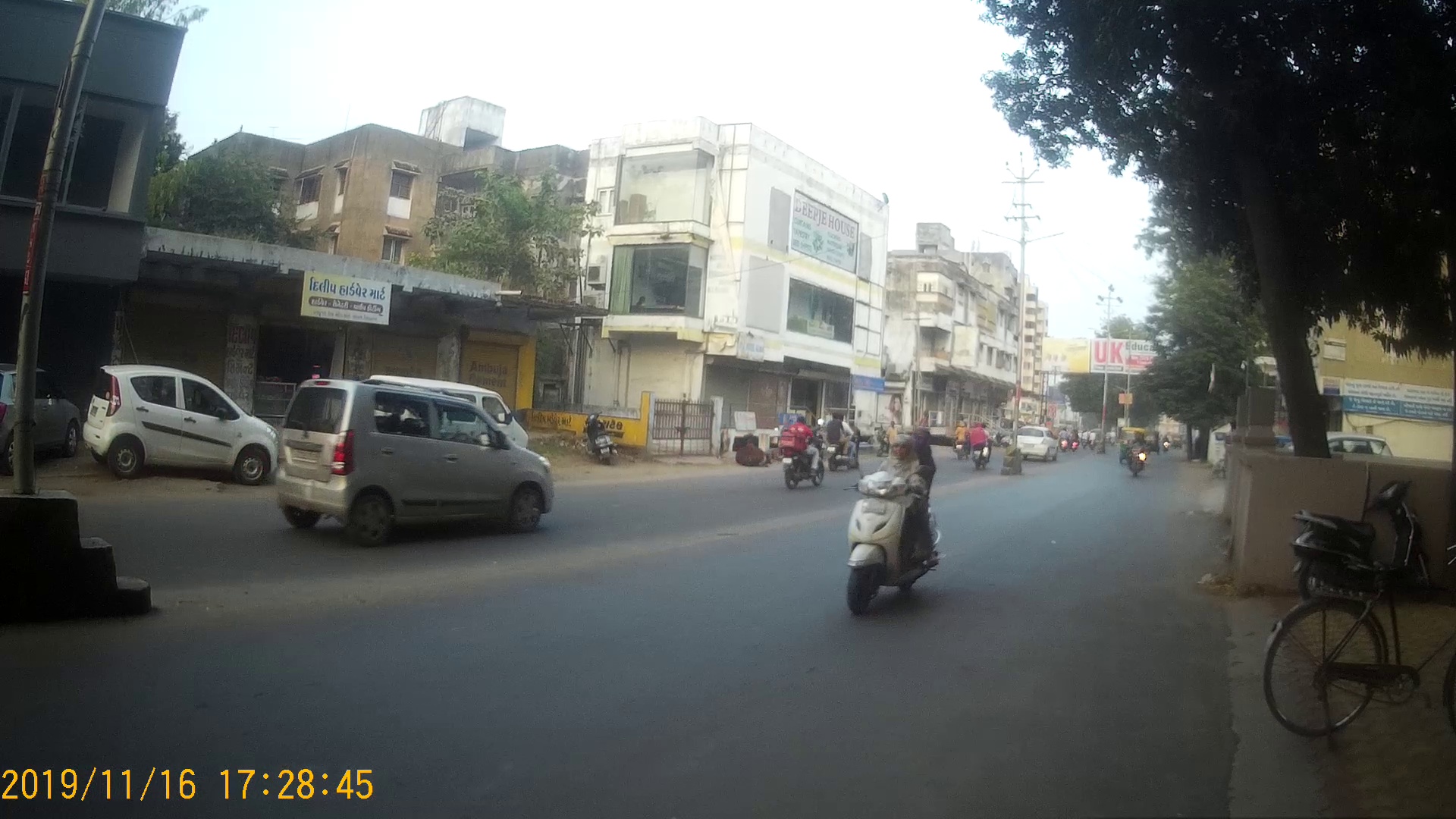}}\hfil
    \subfloat{\includegraphics[width=0.15\textwidth]{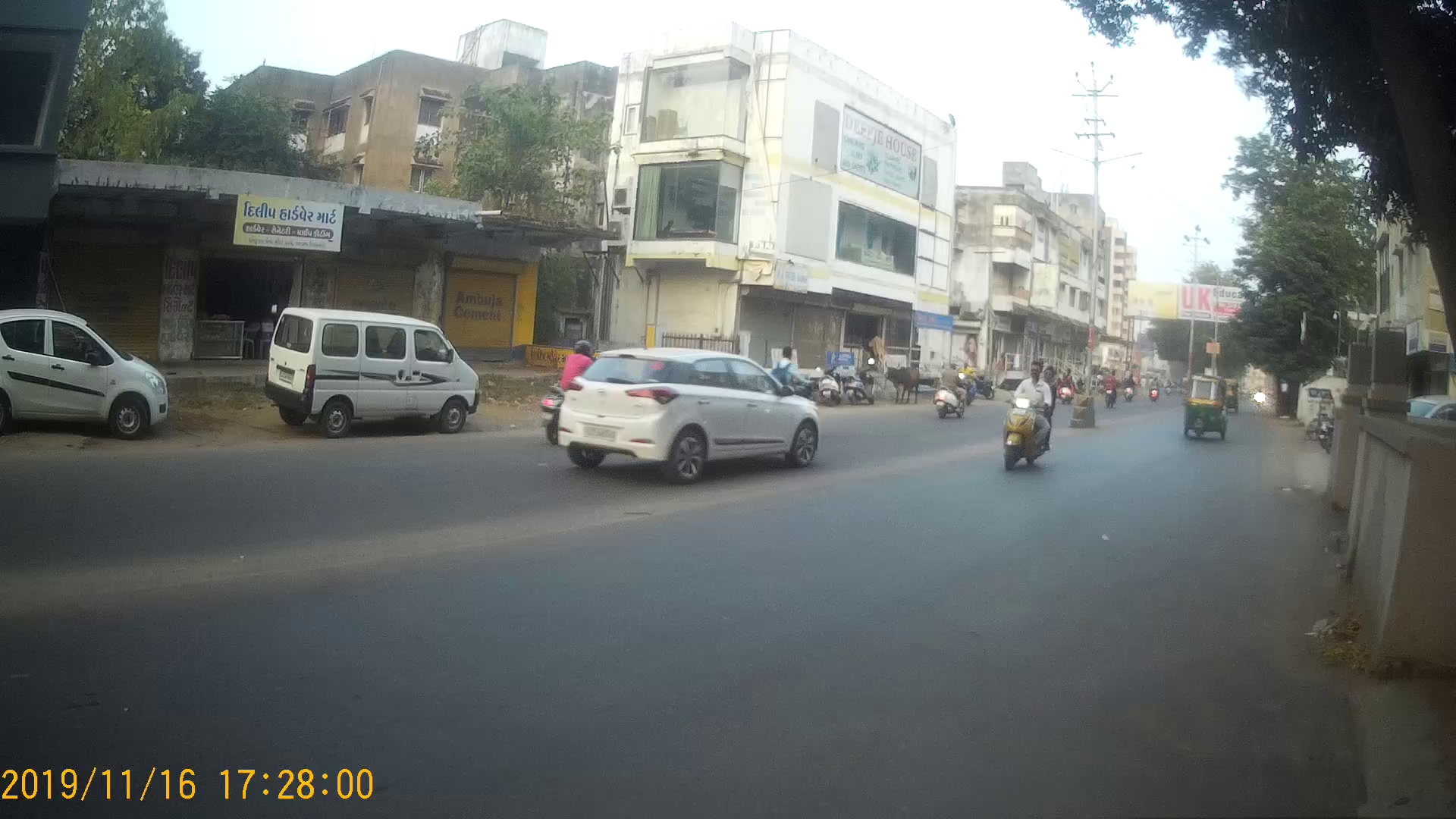}}
\end{figure*}

\begin{figure*}
    \centering
    \subfloat[True positive]{\includegraphics[width=0.15\textwidth]{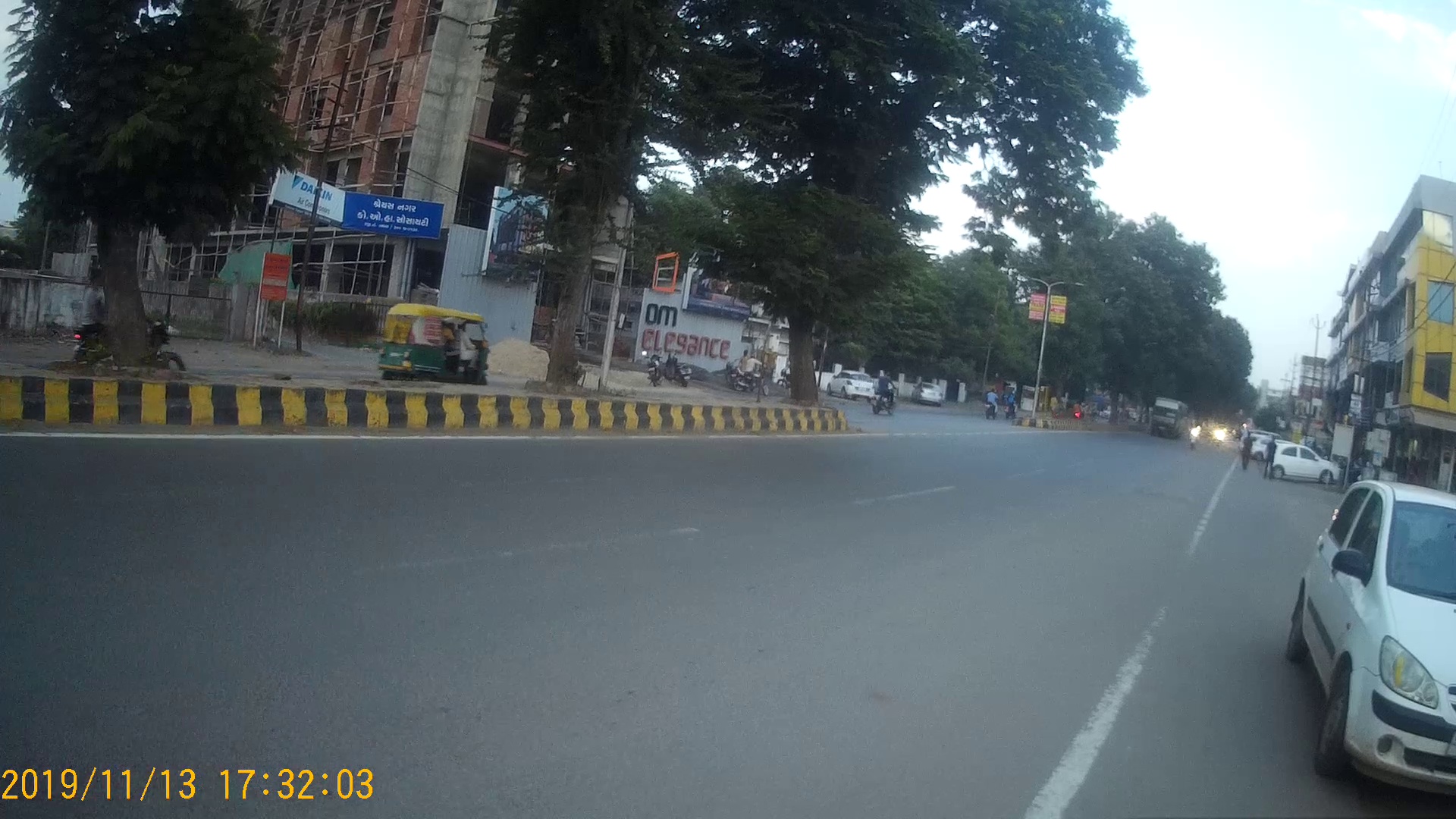}}\hfil
    \subfloat[True positive]{\includegraphics[width=0.15\textwidth]{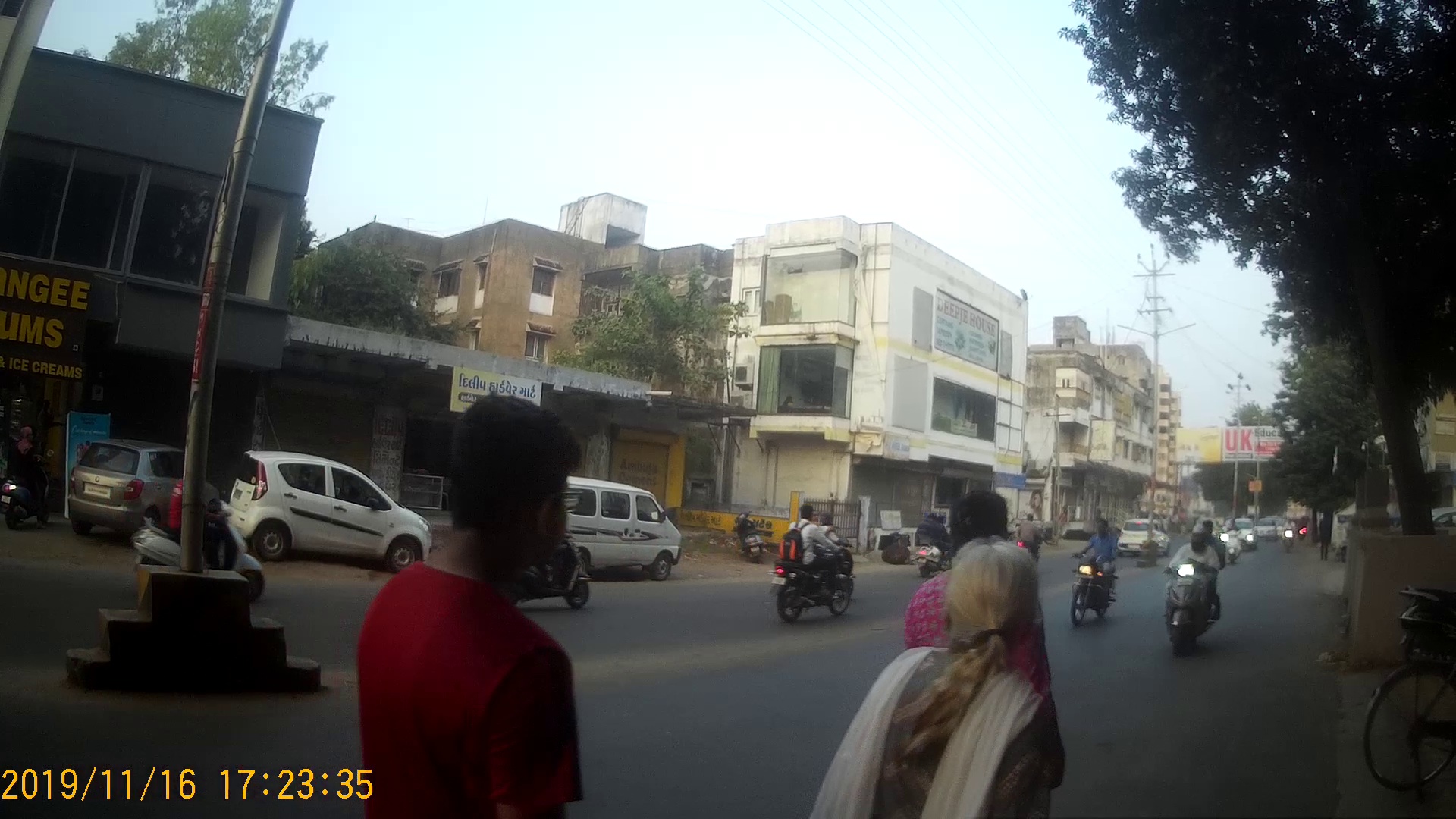}} \hfil
    \subfloat[True negative]{\includegraphics[width=0.15\textwidth]{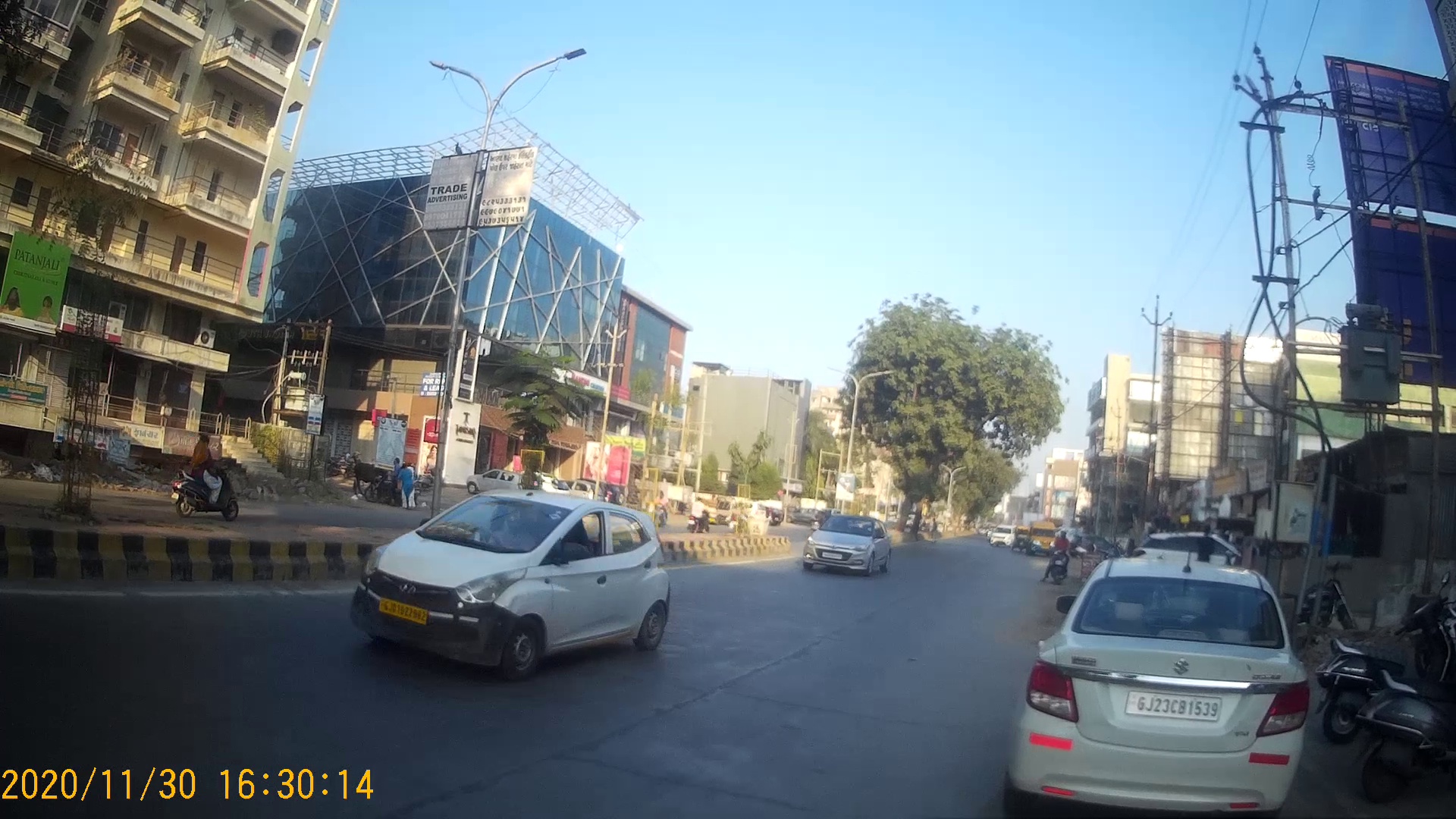}}\hfil
    \subfloat[True negative]{\includegraphics[width=0.15\textwidth]{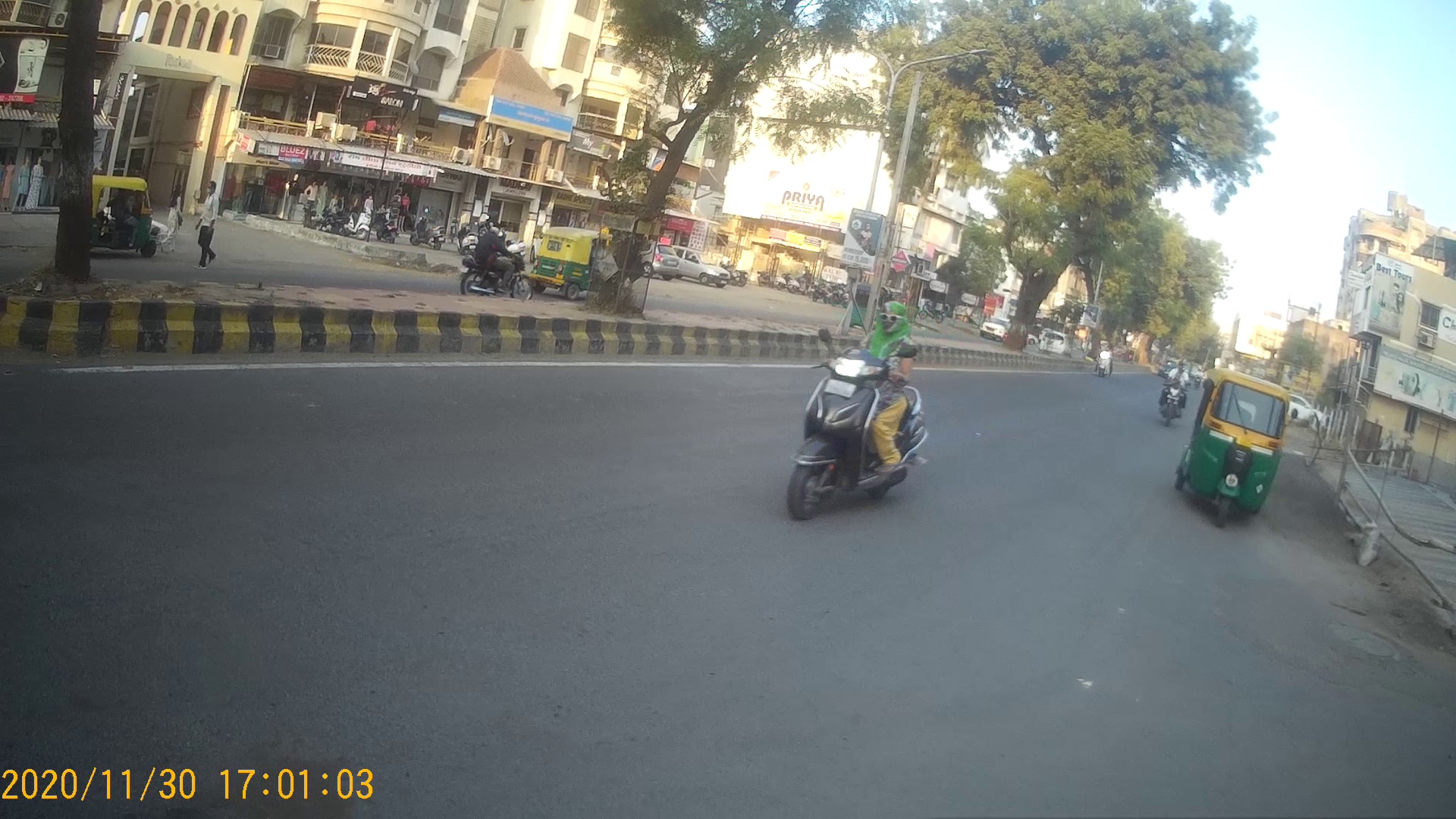}}\hfil
    \subfloat[False positive]{\includegraphics[width=0.15\textwidth]{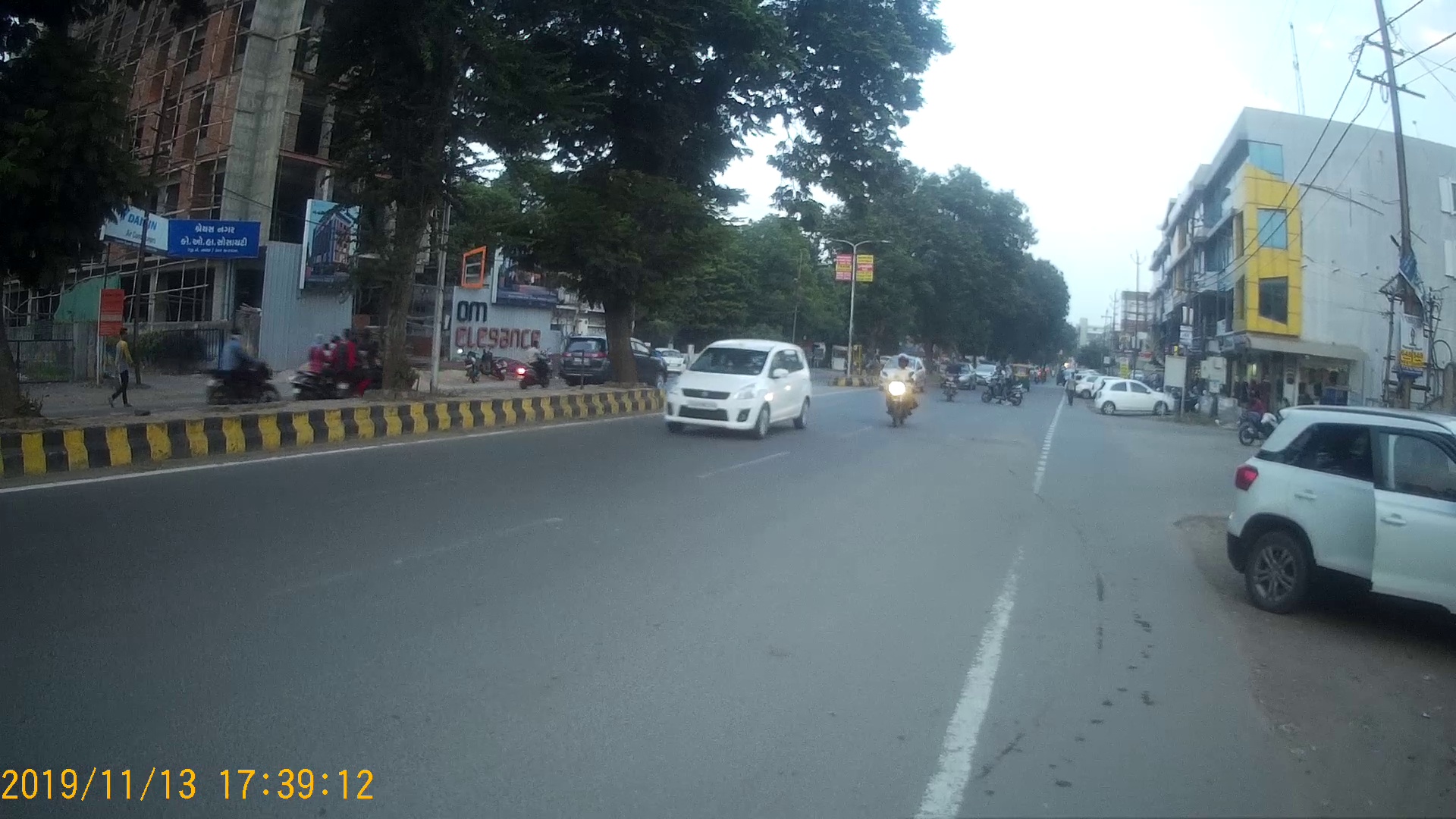}}\hfil
    \subfloat[False negative]{\includegraphics[width=0.15\textwidth]{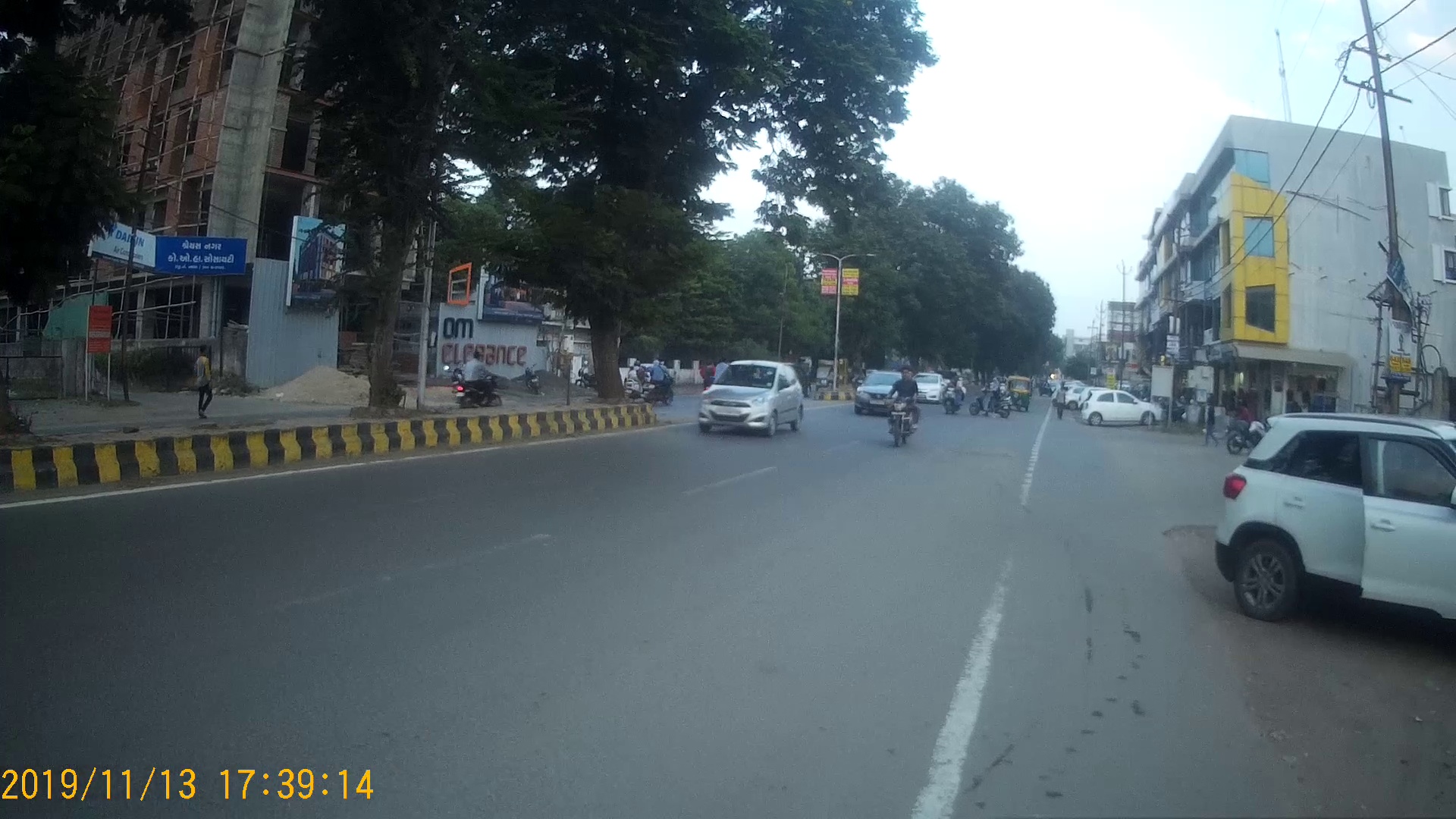}}
    \caption{From top to bottom: sample prediction outputs of single frame SVM and multi frame SVM respectively.}
\end{figure*}

As shown in Figure 6 , we used a sliding window based approach for feature extraction. For predicting label of a particular frame, we considered an entire window of frames (i.e., the current frame plus past k frames). The feature vector for a particular frame consists of 24 features of that frame (using the per frame features explained in single frame SVM section), and it additionally consists of predicted labels for previous k-1 frames using saved model of single frame SVM. So if we take window size = 10, the length of feature vector will be 24 + 9 = 33. We applied minmax scaling to the extracted features and trained an SVM on the train set.

\subsection{Results}

\begin{table}[htp]
  \caption{ML methods: Results on test data}
  \begin{tabular}{ccl}
    \toprule
    Method & Precision & Recall\\
    \midrule
    Single frame SVM & 0.68 & 0.87 \\
    Multi frame SVM & 0.79 & 0.84 \\
  \bottomrule
\end{tabular}
\end{table}
Table 2 shows the precision and recall scores (using threshold of 0.5 to predict the frame as safe/unsafe) on test split of INDRA achieved by the aforementioned ML methods. Figure 8 shows sample prediction outputs for the ML methods. While we feel that it was a good idea to start with simple SVMs as we were working on a completely new dataset (without any prior baseline performance for this task), instead of trying to improve these methods, we switched to more advanced deep learning (DL) methods because of the following reasons:
\begin{enumerate}
    \item The precision scores of 0.68 and 0.79 are very poor given that road crossing is a very sensitive application.
    \item For the training experiments, we saved the vehicle bounding boxes generated using RetinaNet beforehand. The entire inference pipeline starting from detecting vehicles, extracting features and predicting the road crossing safety using pre-trained SVM would have been very slow in real-time, thus making it infeasible for deployment.
\end{enumerate}

\section{Deep Learning Methods}

Deep learning (DL) has shown to learn highly effective features from image and video data, yielding high accuracy in many tasks. Hence, we performed many training experiments using convolutional neural networks (CNNs) on individual video frames of INDRA. We performed these experiments on AWS g4dn.xlarge instance (with Tesla T4 GPU). The data preparation and CNN architecture details are explained in this section.

\subsection{Data Preparation}

Firstly, we performed train-test-validation split on the dataset. We used 66 videos in train set, 22 in test set and 16 in validation set.

Then, we created an input data pipeline using Tensorflow's Dataset API \cite{DataPipeline}. For the same, we stored all the individual frames from the videos in a separate folder (as jpg images) and made a list containing all the filenames. Next, we performed the following steps to create input pipeline:

\begin{enumerate}
    \item Created tensorflow dataset instance from slices of the filenames and labels.
    \item Shuffled the data with a buffer size equal to length of the dataset.
    \item Parsed the images from filename to pixel values (used multiple threads to improve speed). Also, we resized the images to the required input size of the models.
    \item Applied data augmentation by using random brightness, contrast and saturation (used multiple threads to improve the speed).
    \item Batched the images (batch size = 16).
    \item Prefetched one batch to make sure that a batch is ready to be served at all time.
\end{enumerate}

\subsection{Class weights}

Using class weights was essential during model training since the dataset is highly unbalanced (obviously there are far more unsafe frames than safe frames in traffic scenarios). Hence we used class weights as: 1 for unsafe class, 1.92 for safe class in each of the following DL methods. We obtained the value of 1.92 by calculating the ratio of number of unsafe frames to the number of safe frames from train set.

\subsection{Baseline Model - MobileNetV2}

As a baseline model, we used the MobileNetV2 architecture \cite{sandler2018mobilenetv2} with additional dense layers at the top. We used MobileNetV2 architecture because it is very memory efficient and lightweight, making it highly efficient for mobile and embedded vision applications. The key properties of MobileNetV2 architecture that makes it lightweight and efficient are the use of depthwise separable convolutions and an inverted residual structure. 

The model architecture is as shown in Figure 9. On top of the last convolutional layer of MobileNetV2, we added an average pooling layer for downsampling, followed by a dropout layer and a fully connected layer with 1 unit (i.e., the output layer giving the probability of a frame being safe).

\begin{figure}[htp]
            
            \includegraphics[width=\columnwidth]{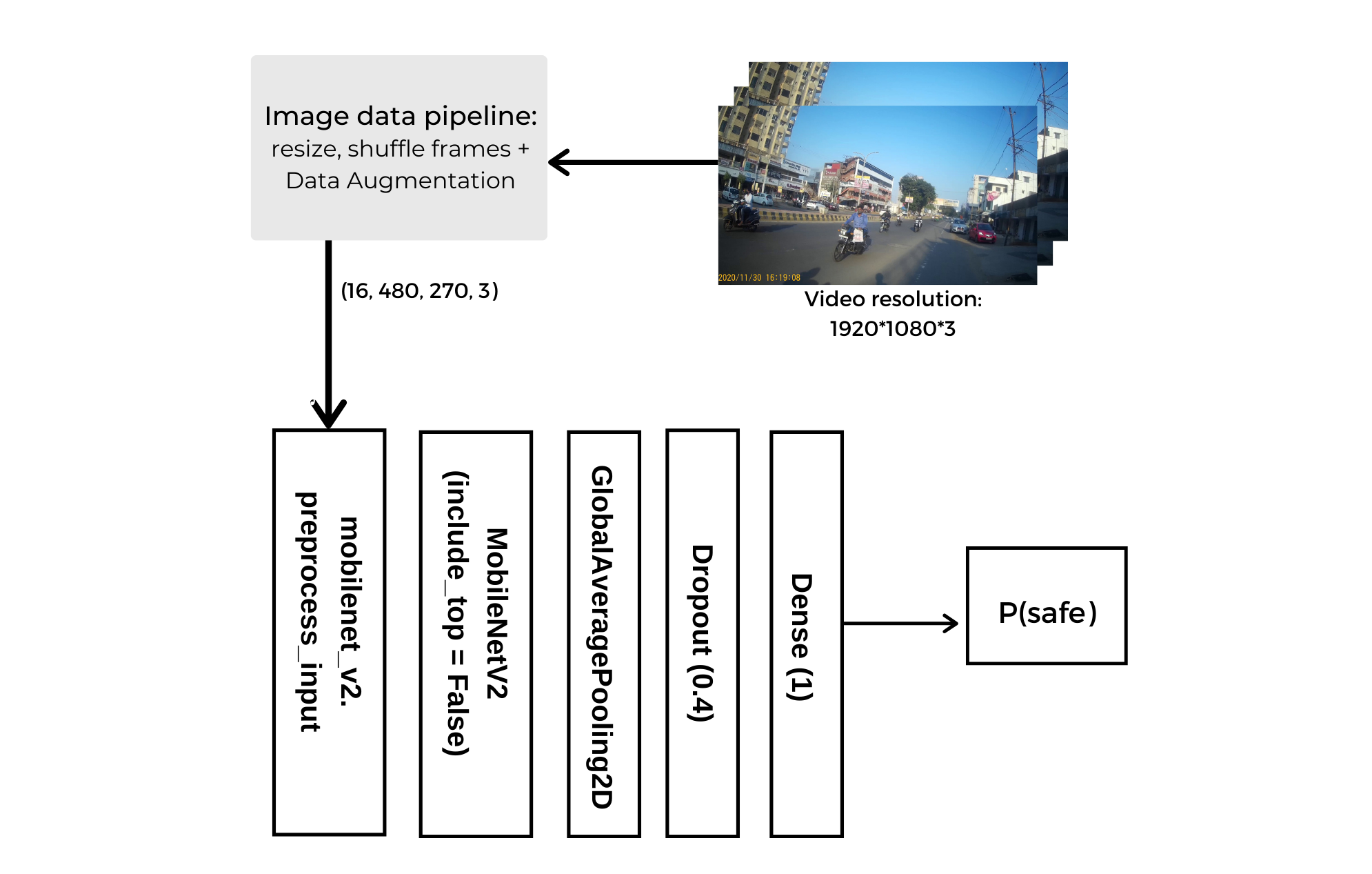}
            \caption{Baseline Model - MobileNetV2}
\end{figure}

We trained the model for 150 epochs minimizing the binary cross entropy loss. We used dropout rate of 0.4 for the additional layer, and used Adam's optimizer with learning rate of 5e-4.

\subsection{RoadCrossNet}

\begin{figure}[htp]
            \centering
            \includegraphics[width=\columnwidth]{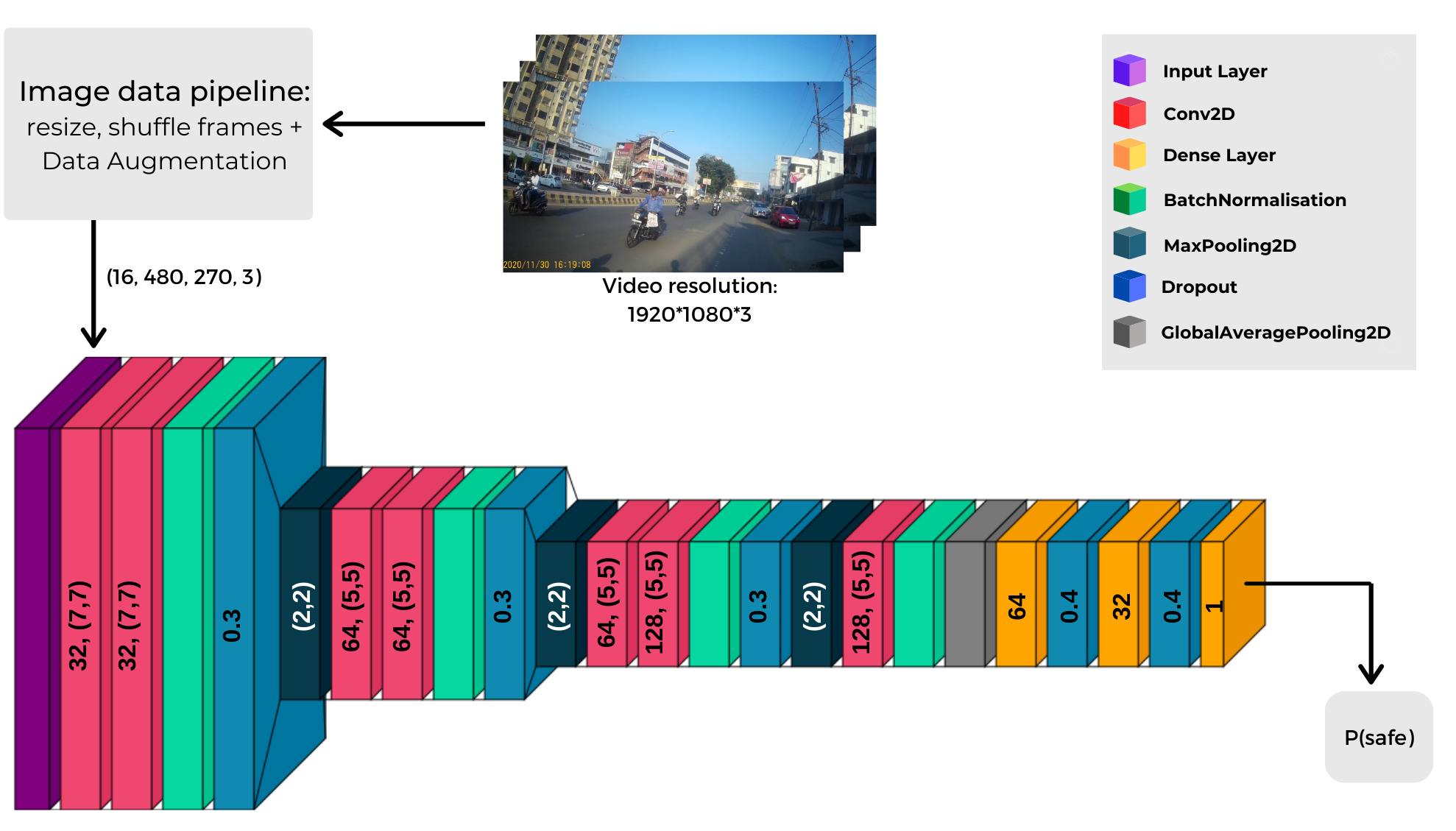}
            \caption{RoadCrossNet}
\end{figure}

We observed that MobileNetV2 exhibited a clear case of overfitting, so we attempted to develop a new lightweight architecture: \textbf{\textit{RoadCrossNet}} from scratch. The model architecture is as shown in Figure 10.

Similar to the trend in many neural network architectures, in RoadCrossNet, we increased the number of filters and decreased the filter size while going deeper in the network. As shown in the architecture diagram, we also used multiple batch-normalization and dropout layers to prevent overfitting. Aditionally, for downsampling, we used max pooling layers.

We trained the model for 150 epochs minimizing binary cross entropy loss. We used dropout rate of 0.3 for first three dropout layers and used dropout rate of 0.4 for subsequent dropout layers. Also, we used Adam's optimizer with learning rate of 5e-4 for training, and used ReLU activation for the layers.

\subsection{DilatedRoadCrossNet}

\begin{figure}[htp]
            \centering
            \includegraphics[width=\columnwidth]{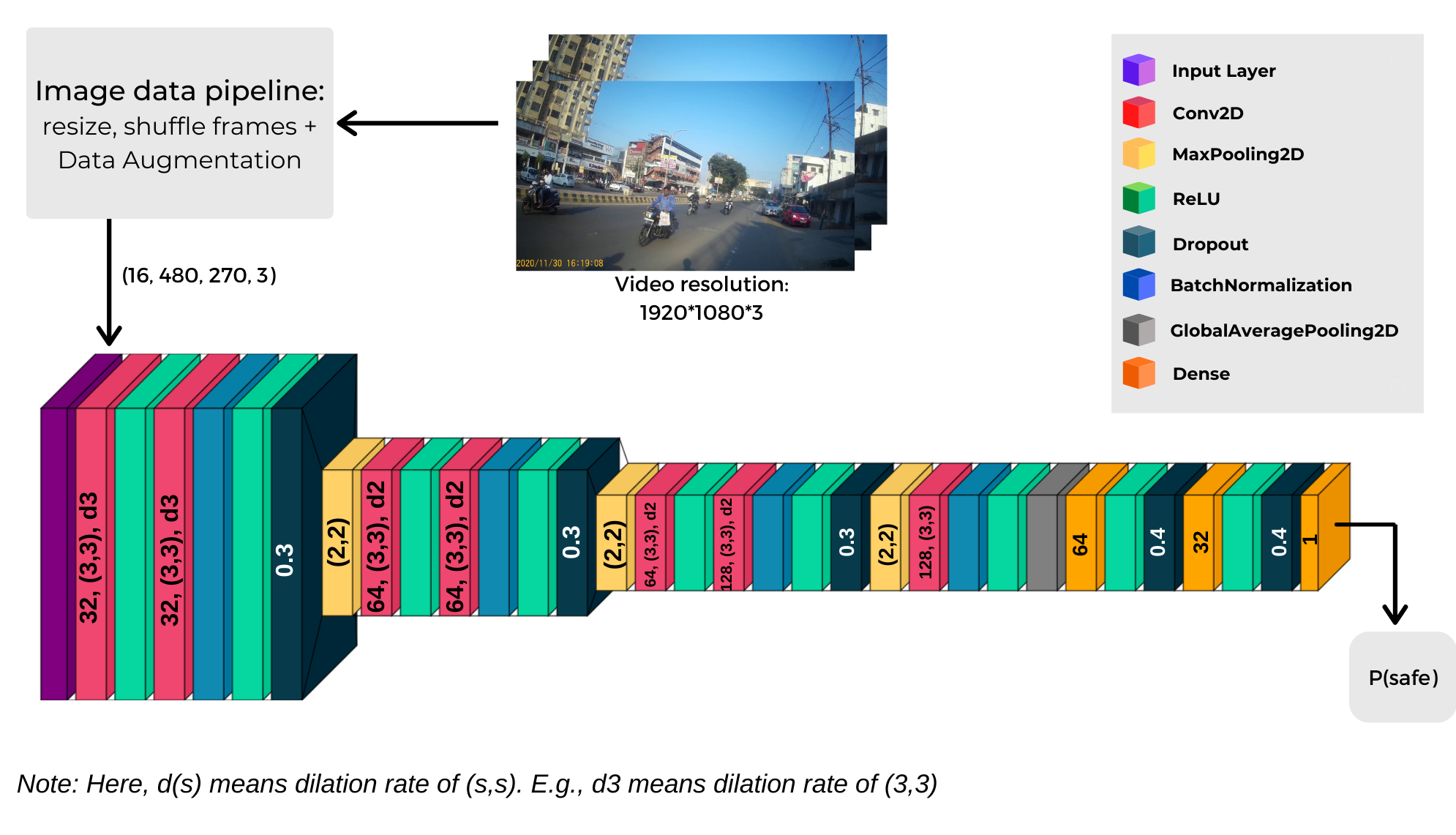}
            \caption{DilatedRoadCrossNet}
\end{figure}

Since it is difficult to optimize convolution layers with larger kernel size (optimization required for real-time deployment), we improved the RoadCrossNet architecture by reducing the kernel sizes of initial layers (which were 7*7 and 5*5). Also, to compensate for the reduced receptive field caused by decrease in kernel size, we used dilated convolutions instead of simple convolutions \cite{luo2016understanding}. Intuitively speaking, receptive field plays an important role in this application since for crossing roads, one needs to consider his/her entire field of vision to make the decision, and not just a particular region from the field of vision. The model architecture is as shown in Figure 11. Apart from using dilated convolutions, the architecture is the same as RoadCrossNet (containing batch normalization, dropout and max pooling layers, with 3 dense layers at the top).

We trained the model for 150 epochs minimizing binary cross entropy loss. We used dropout rate of 0.3 in first three dropout layers and used dropout rate of 0.4 in later two layers. Also, we used Adam's optimizer with learning rate of 5e-4 for training. 
\begin{figure*}
    \centering
    \subfloat{\includegraphics[width=0.15\textwidth]{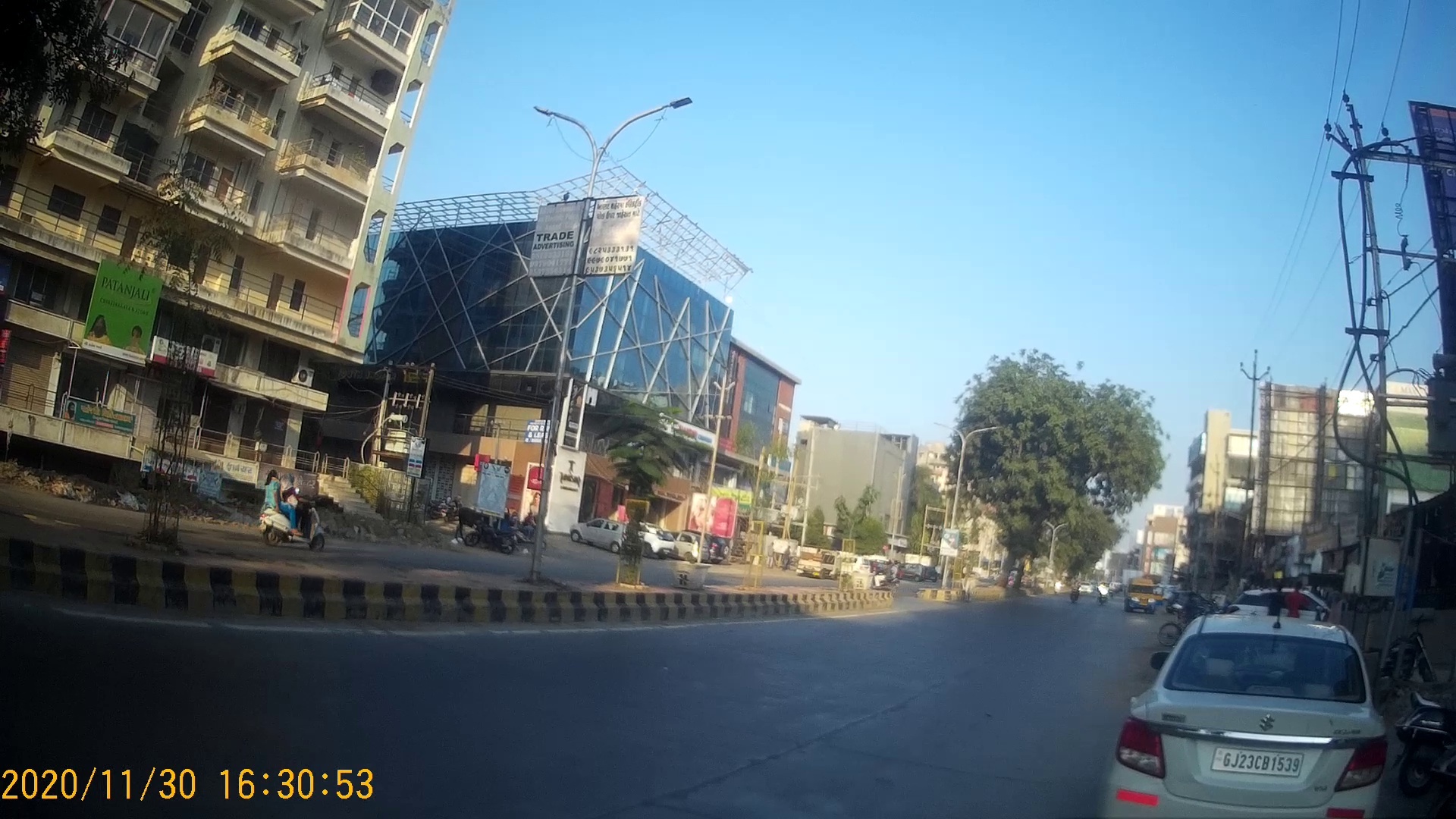}}\hfil
    \subfloat{\includegraphics[width=0.15\textwidth]{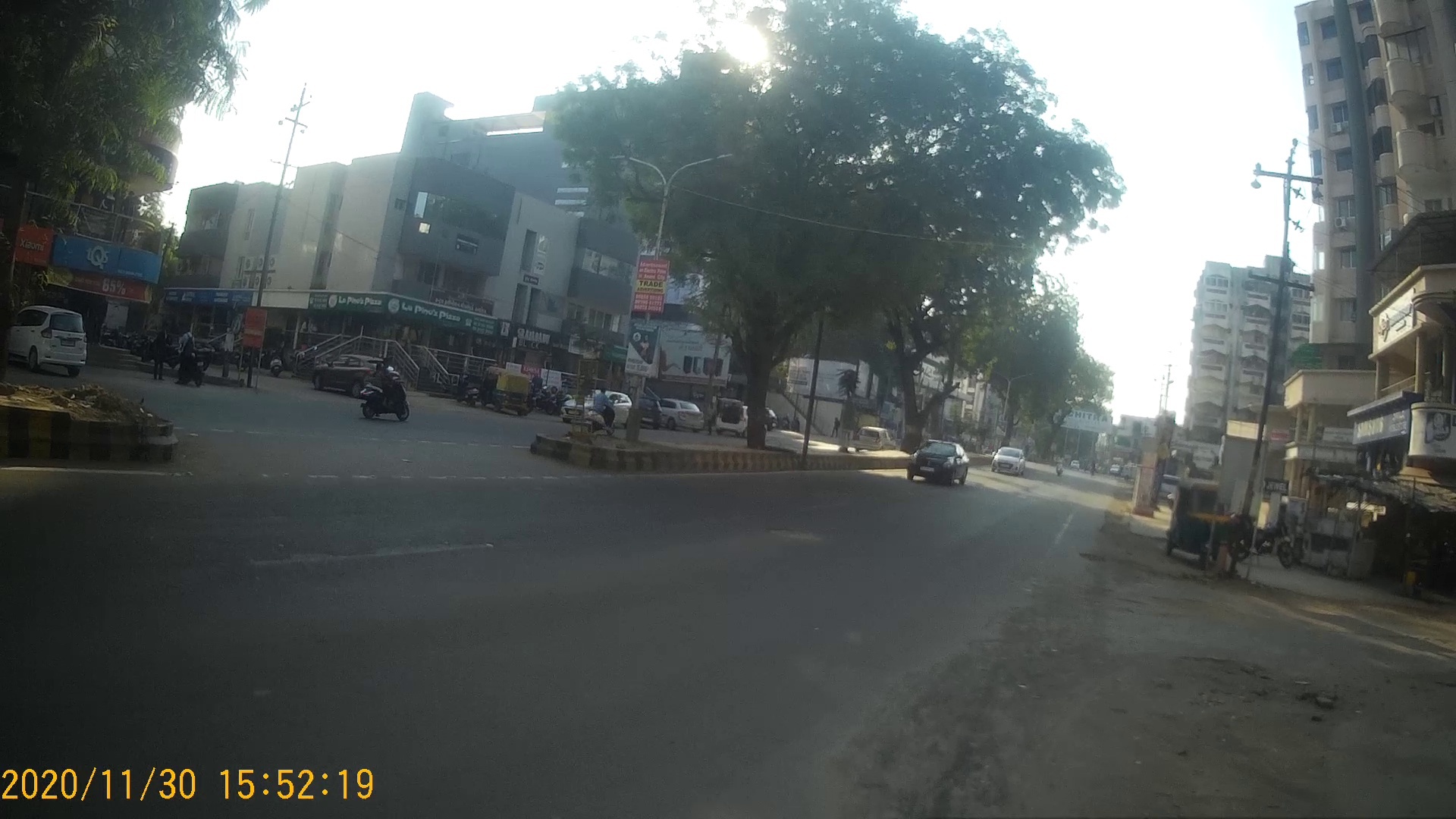}} \hfil
    \subfloat{\includegraphics[width=0.15\textwidth]{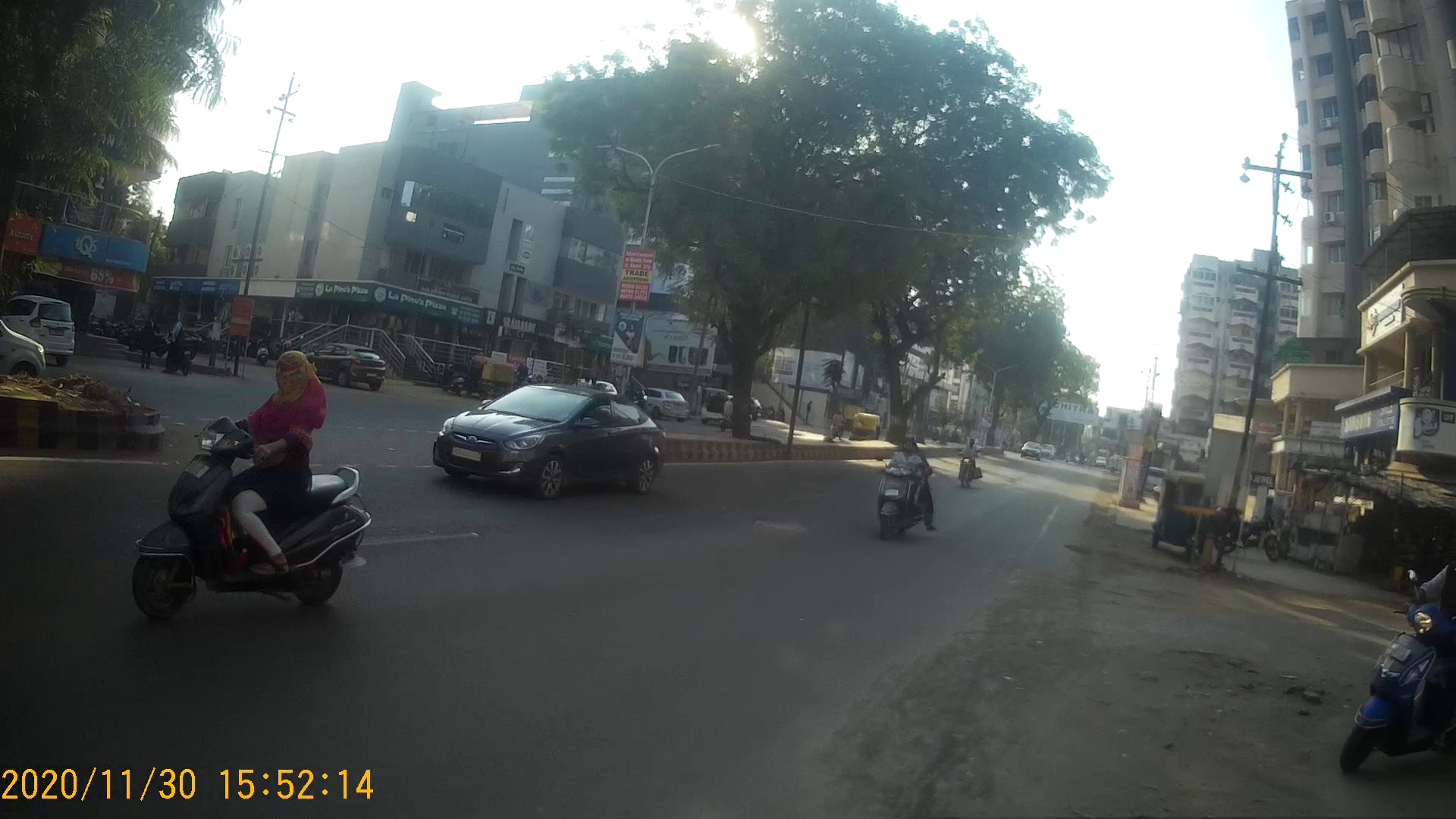}}\hfil
    \subfloat{\includegraphics[width=0.15\textwidth]{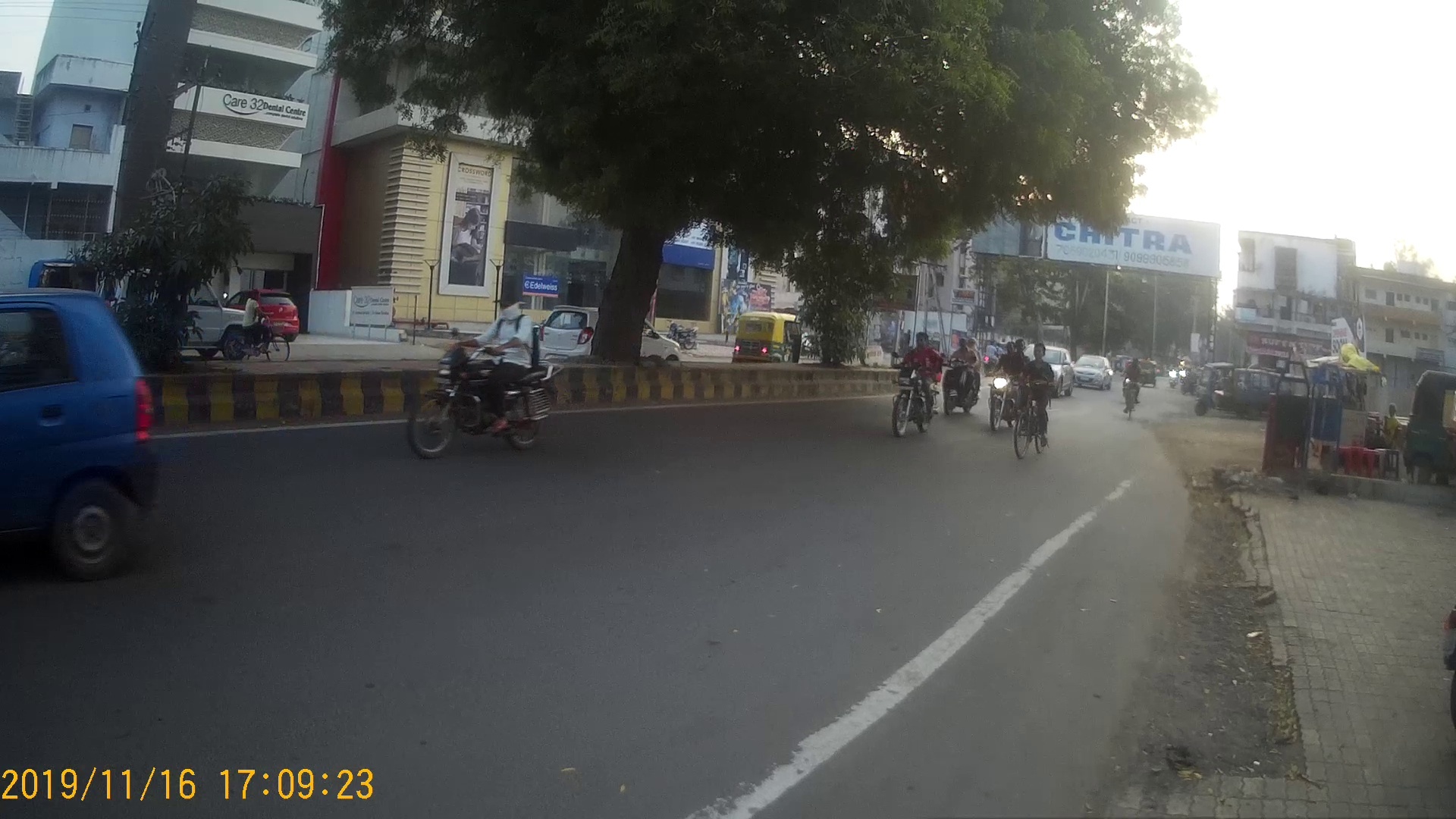}}\hfil
    \subfloat{\includegraphics[width=0.15\textwidth]{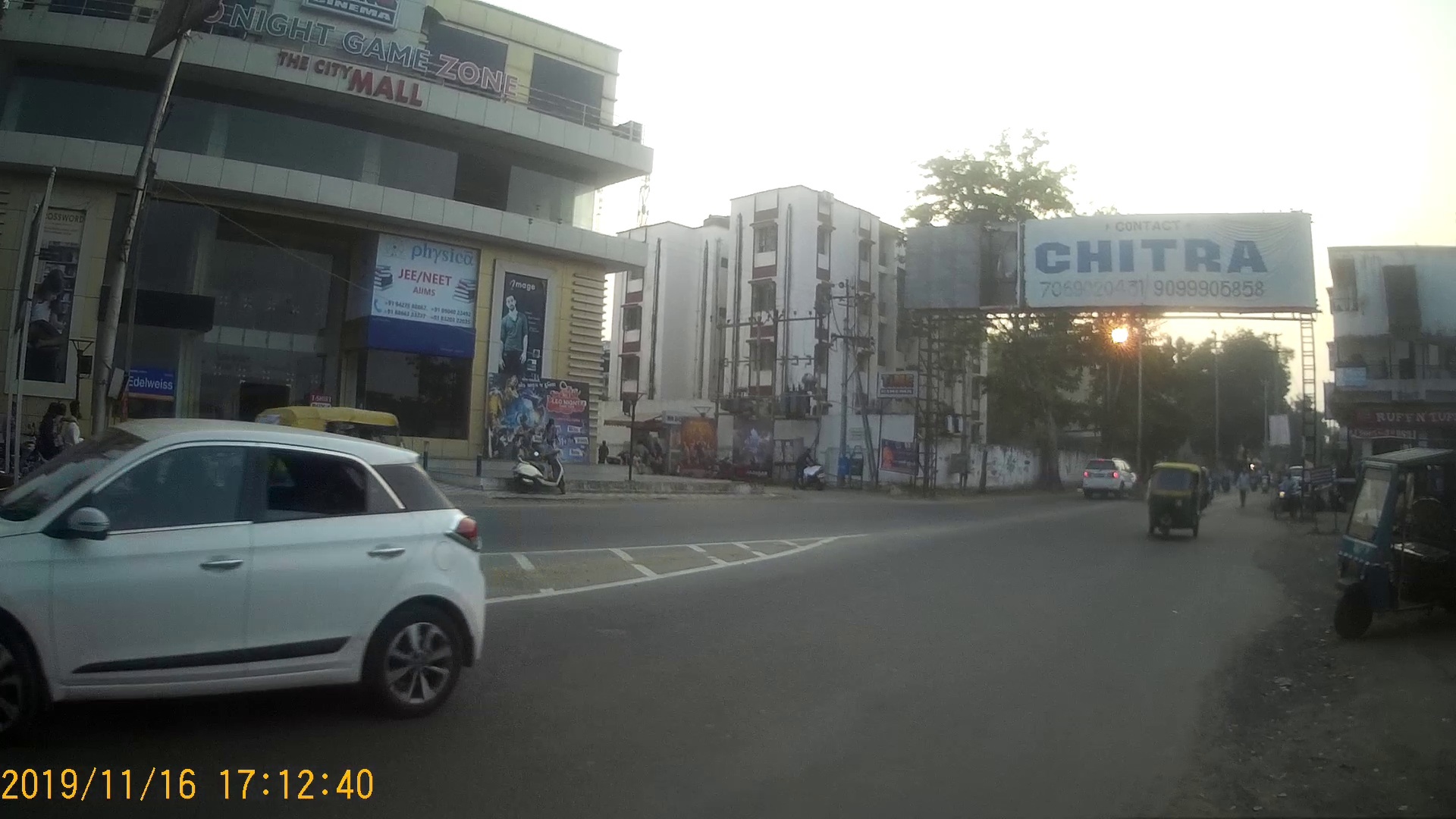}}\hfil
    \subfloat{\includegraphics[width=0.15\textwidth]{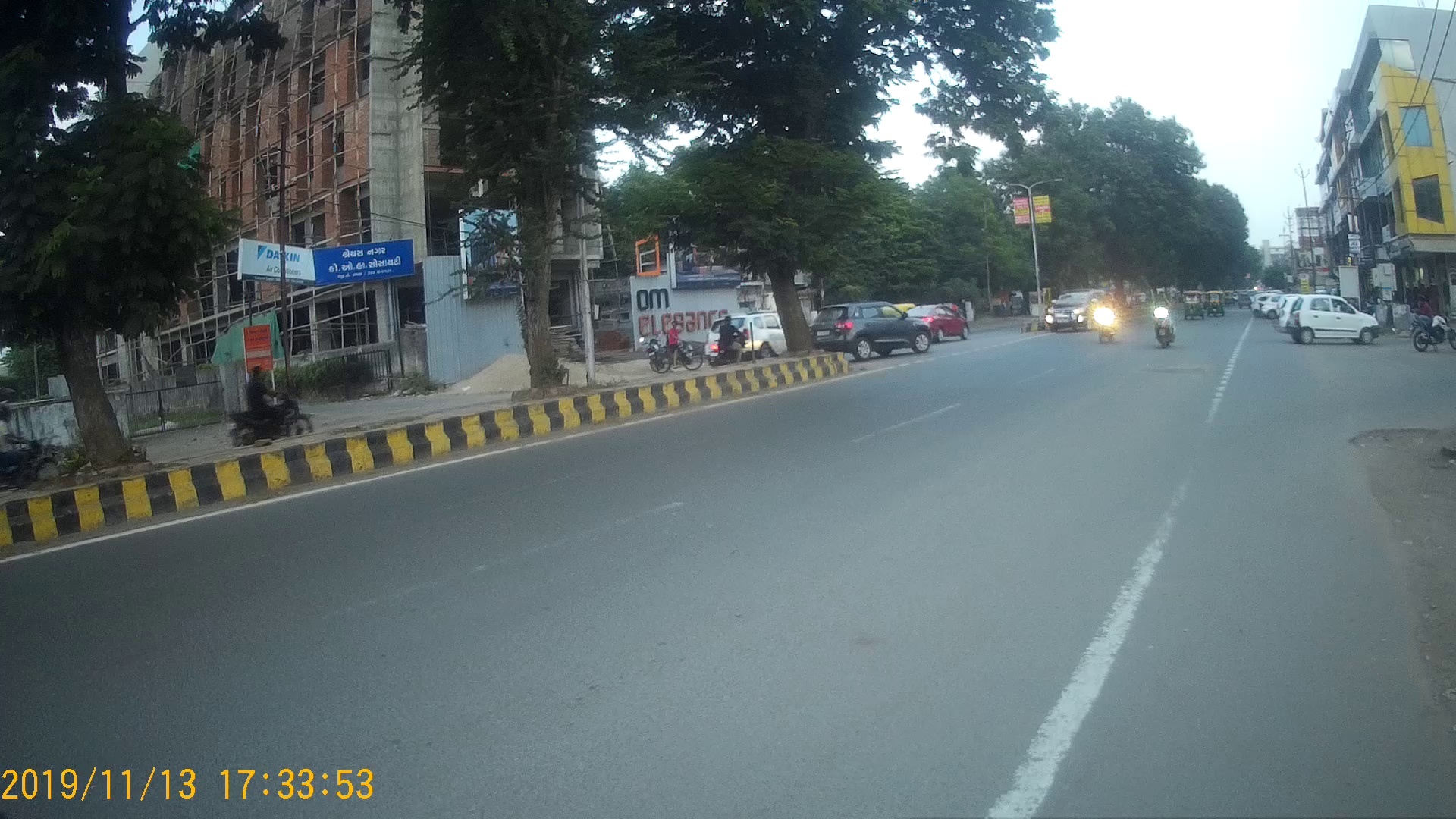}}
\end{figure*}

\begin{figure*}
    \centering
    \subfloat{\includegraphics[width=0.15\textwidth]{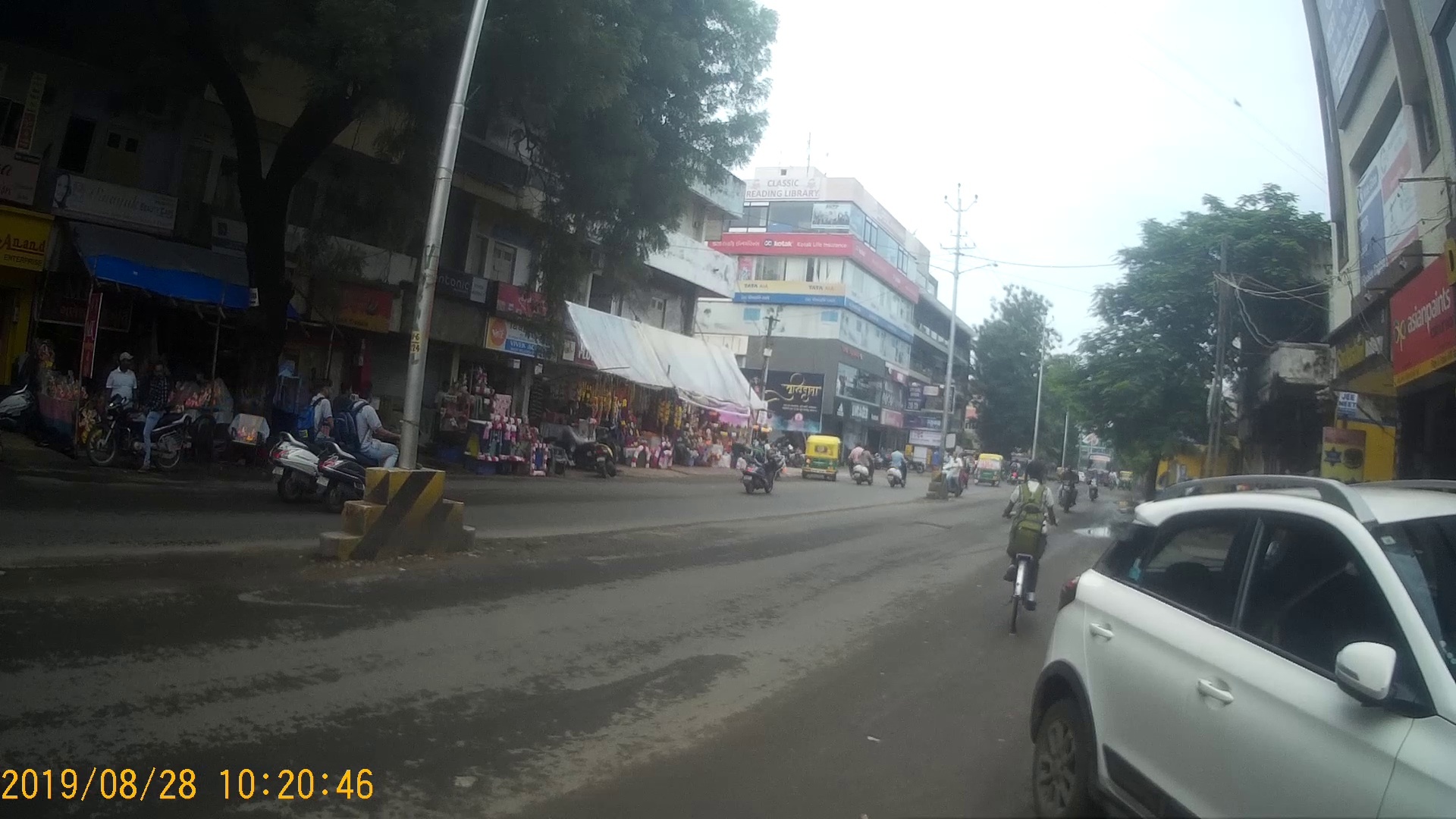}}\hfil
    \subfloat{\includegraphics[width=0.15\textwidth]{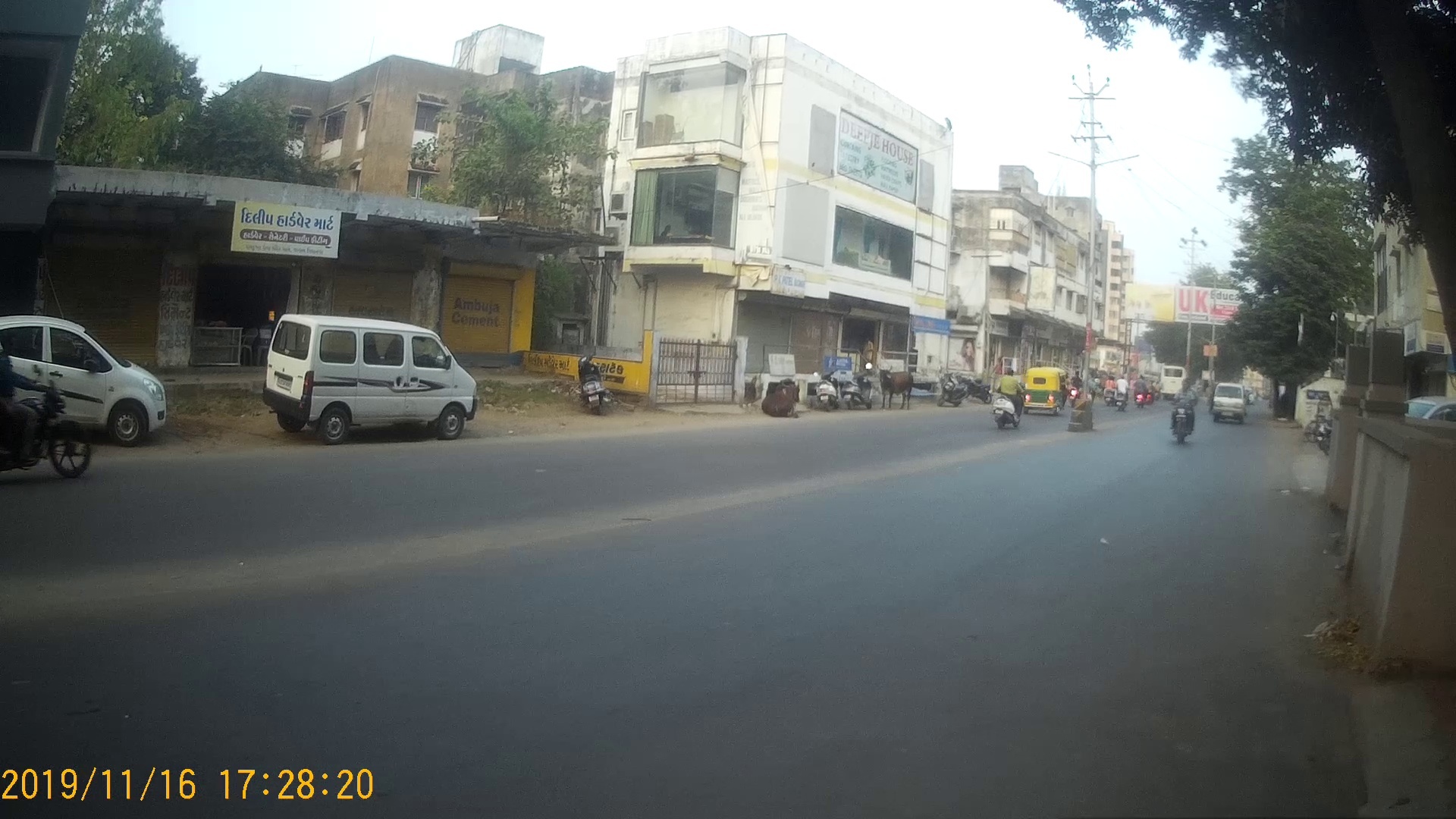}} \hfil
    \subfloat{\includegraphics[width=0.15\textwidth]{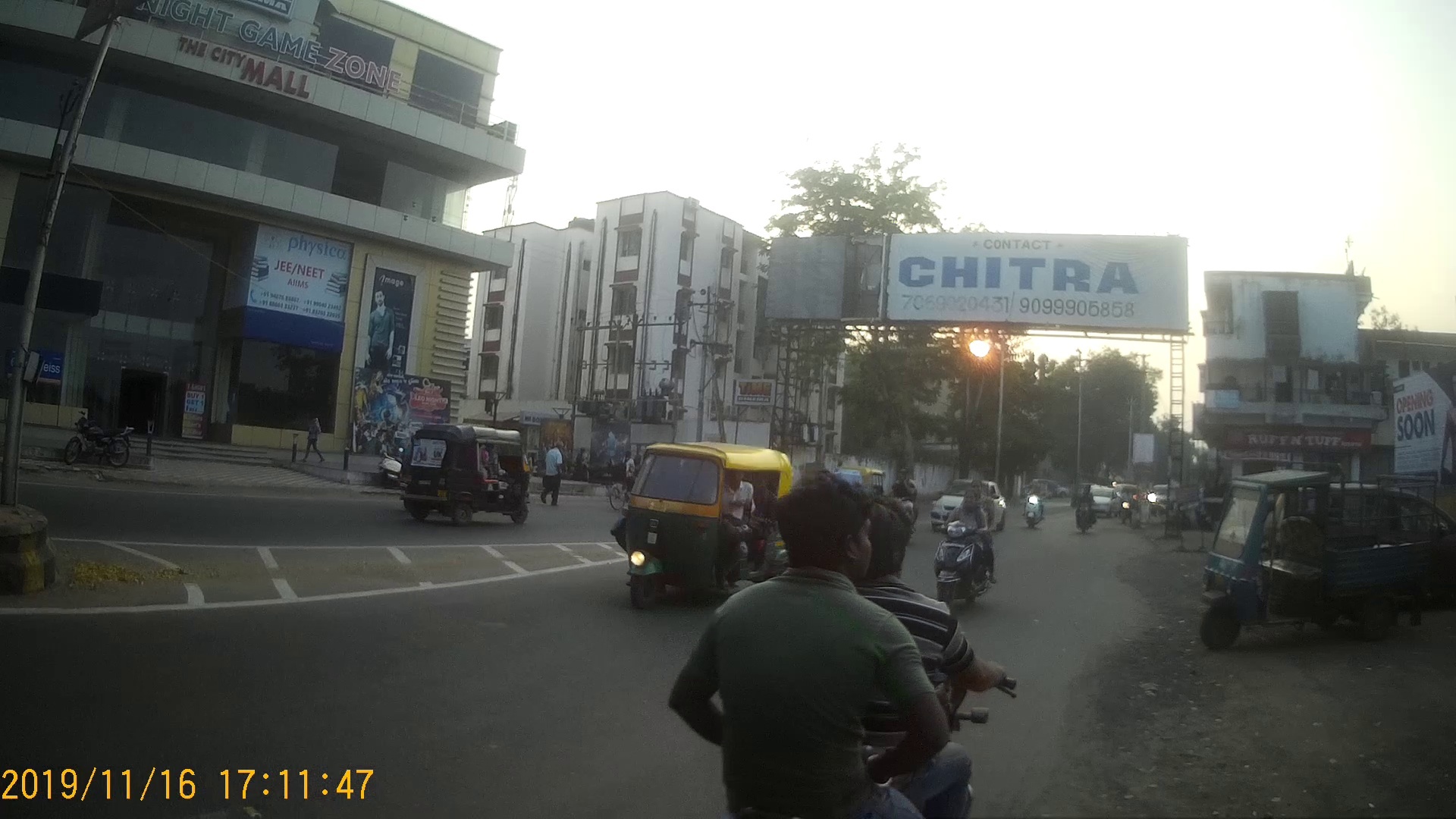}}\hfil
    \subfloat{\includegraphics[width=0.15\textwidth]{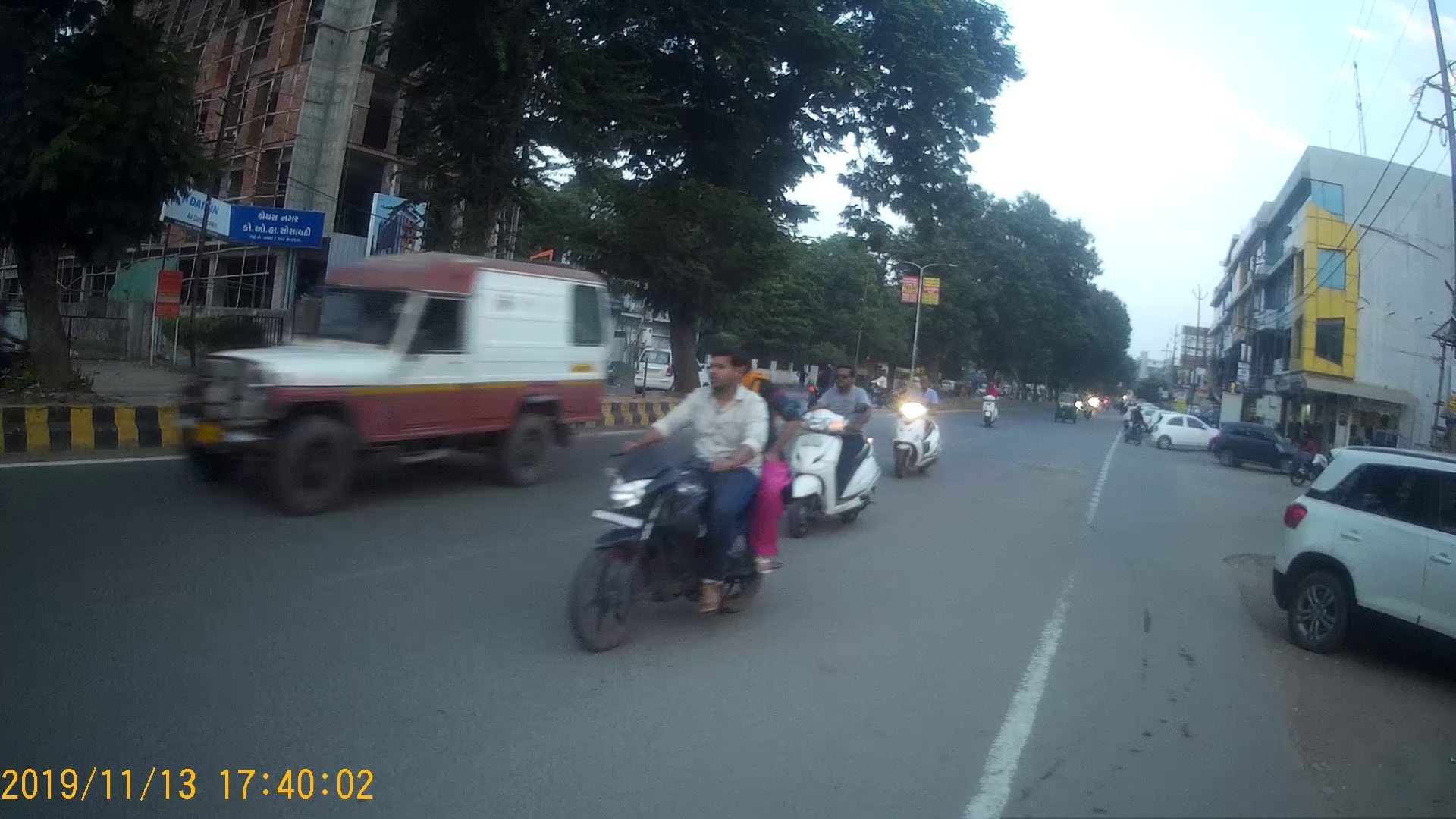}}\hfil
    \subfloat{\includegraphics[width=0.15\textwidth]{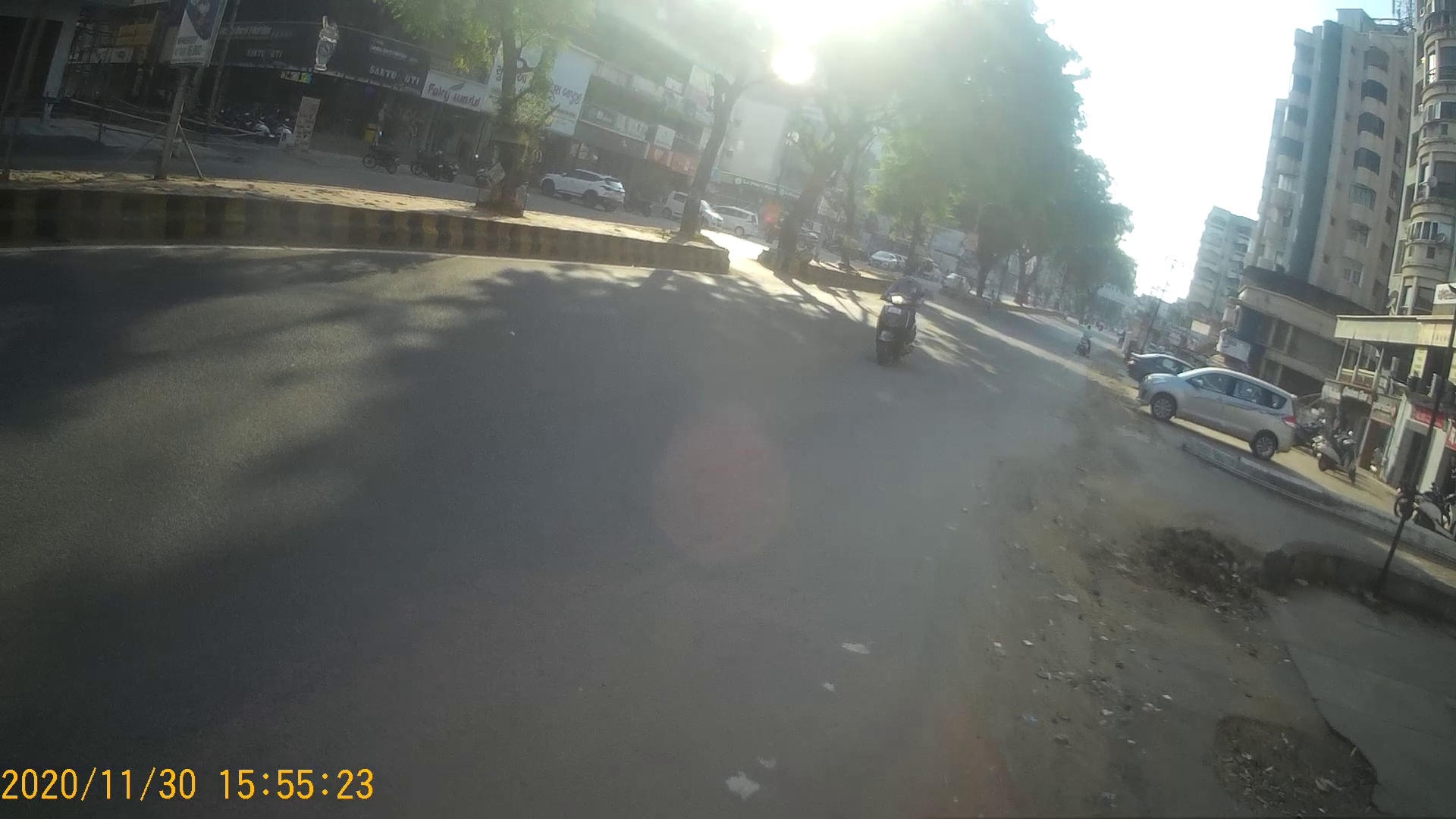}}\hfil
    \subfloat{\includegraphics[width=0.15\textwidth]{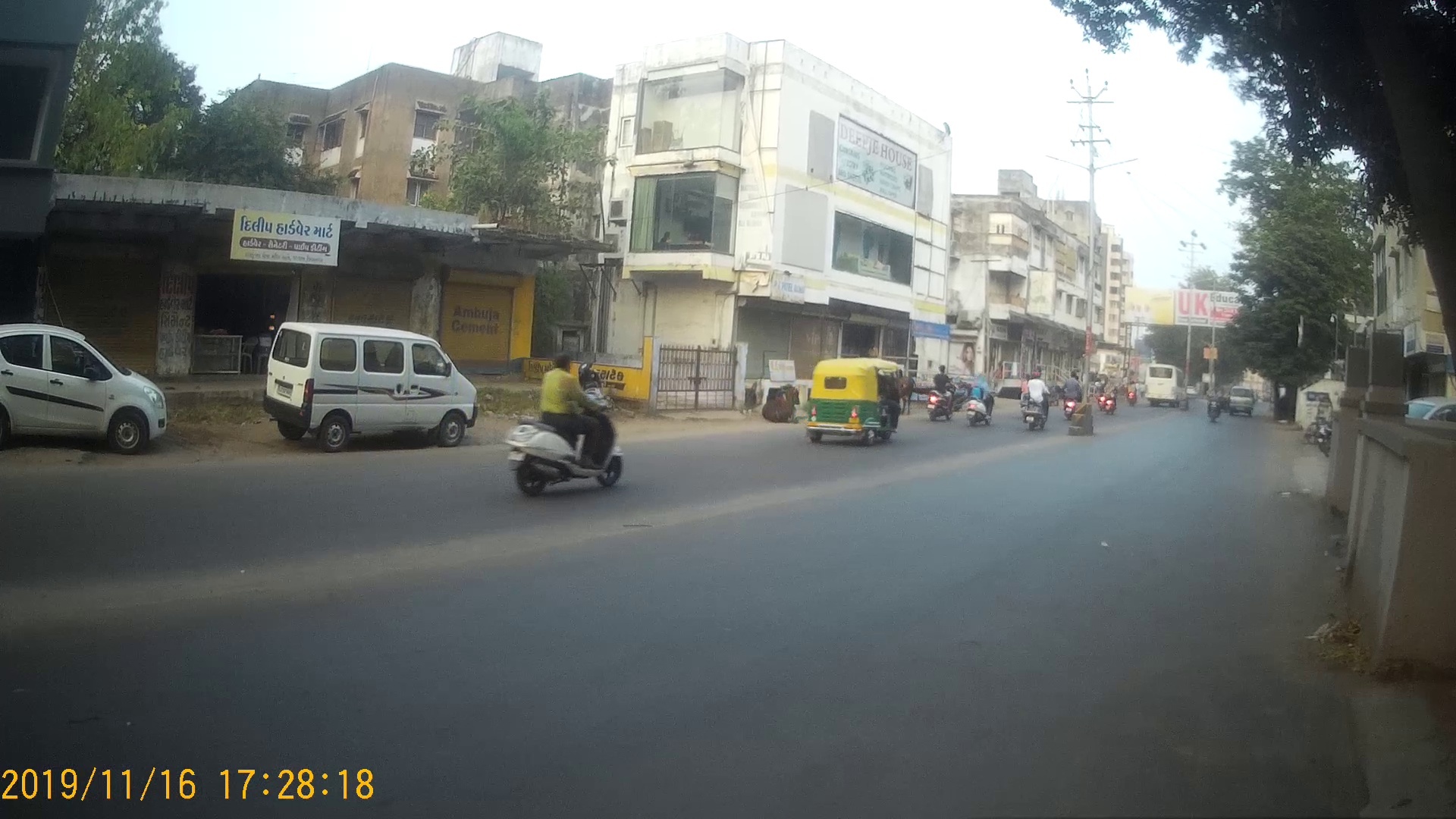}}
\end{figure*}

\begin{figure*}
    \centering
    \subfloat[True positive]{\includegraphics[width=0.15\textwidth]{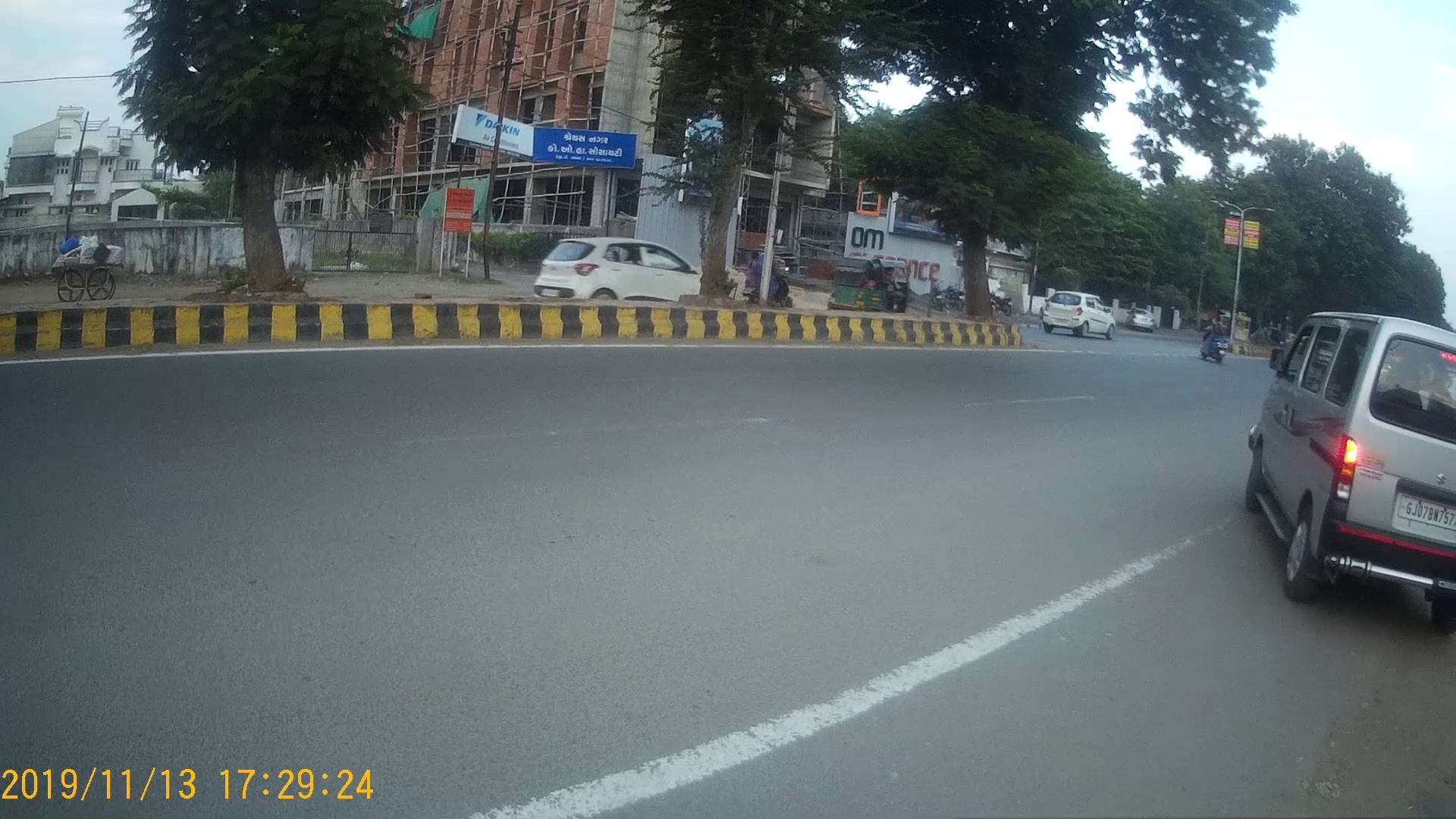}}\hfil
    \subfloat[True positive]{\includegraphics[width=0.15\textwidth]{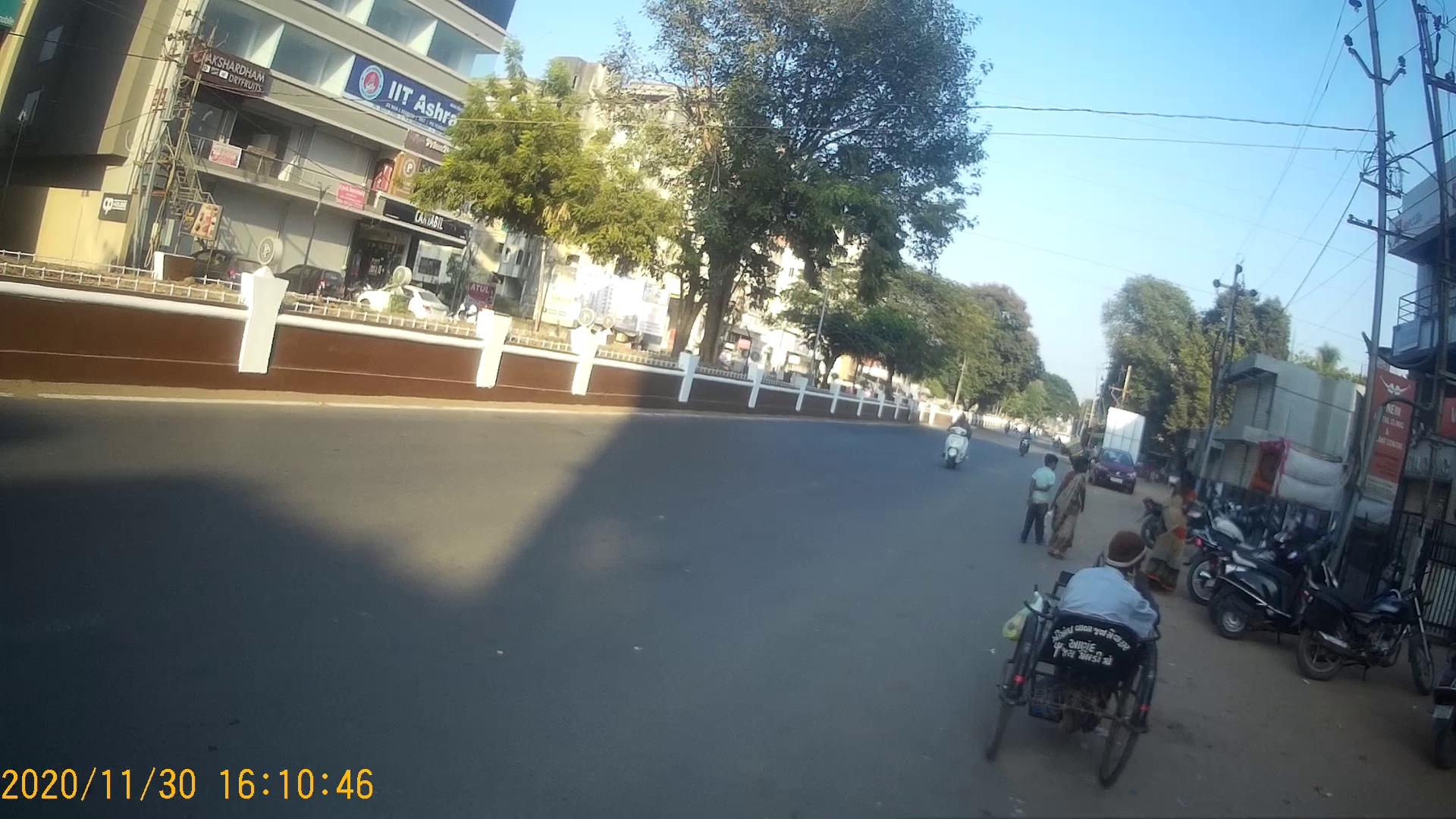}} \hfil
    \subfloat[True negative]{\includegraphics[width=0.15\textwidth]{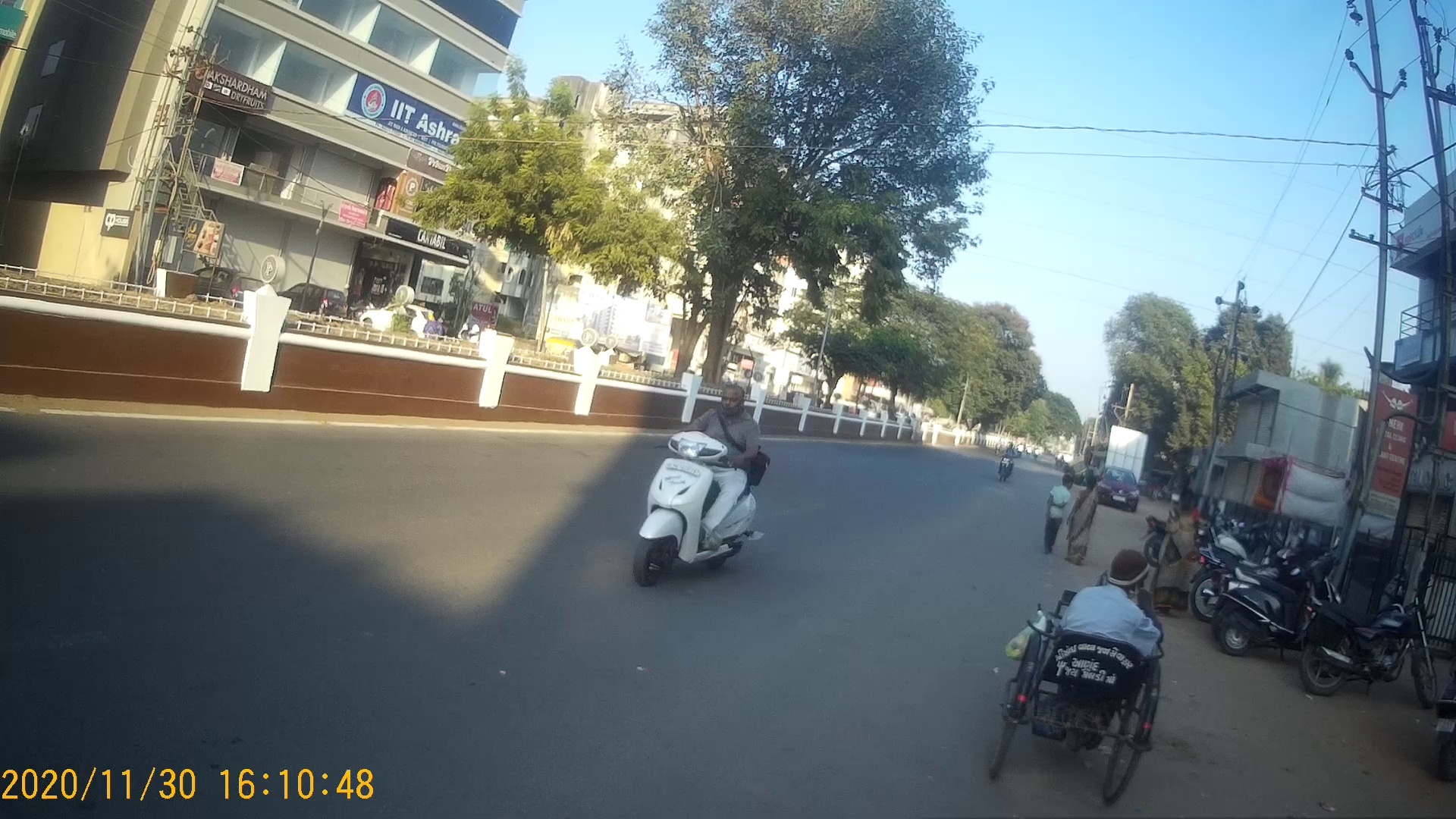}}\hfil
    \subfloat[True negative]{\includegraphics[width=0.15\textwidth]{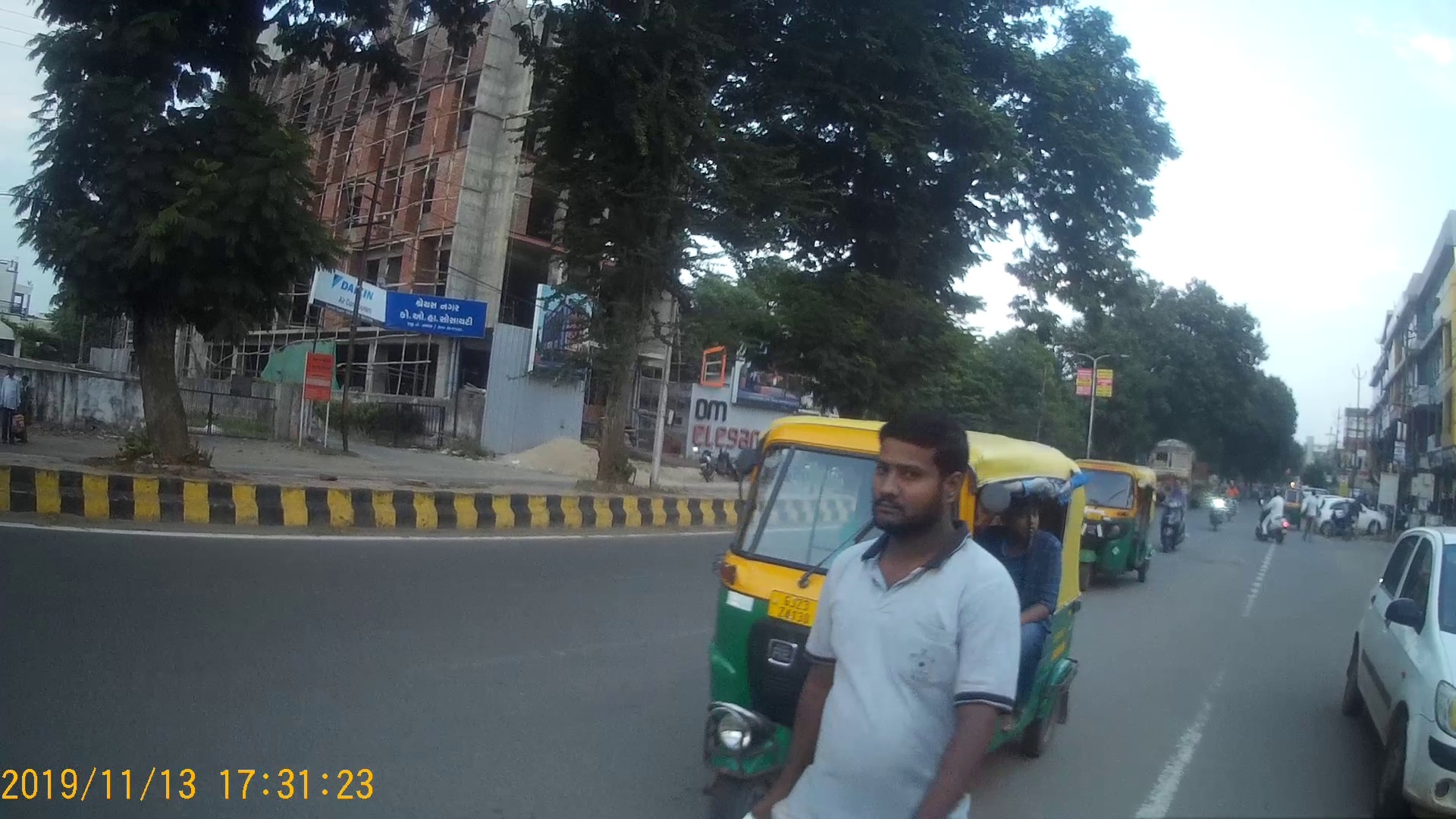}}\hfil
    \subfloat[False positive]{\includegraphics[width=0.15\textwidth]{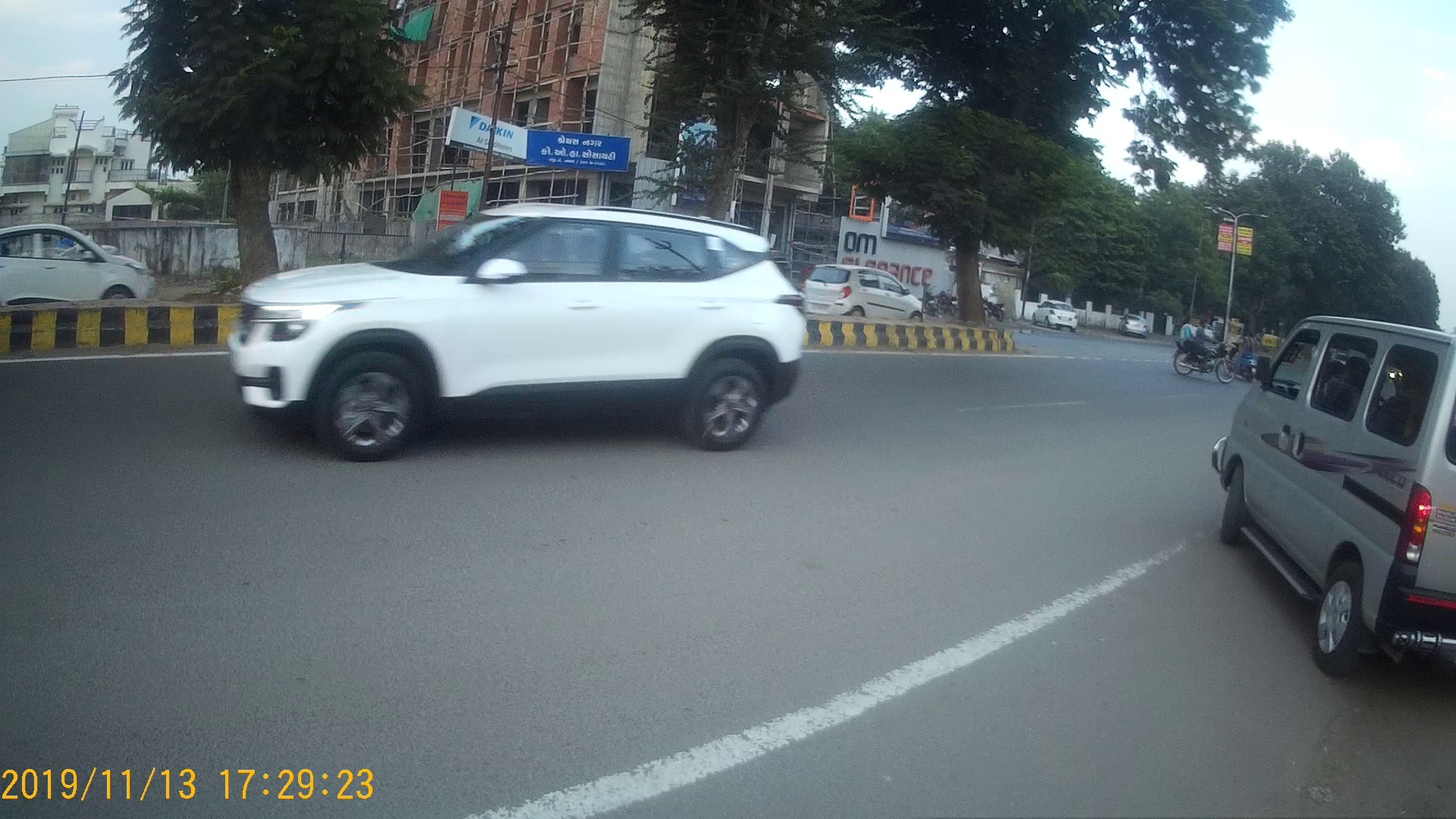}}\hfil
    \subfloat[False negative]{\includegraphics[width=0.15\textwidth]{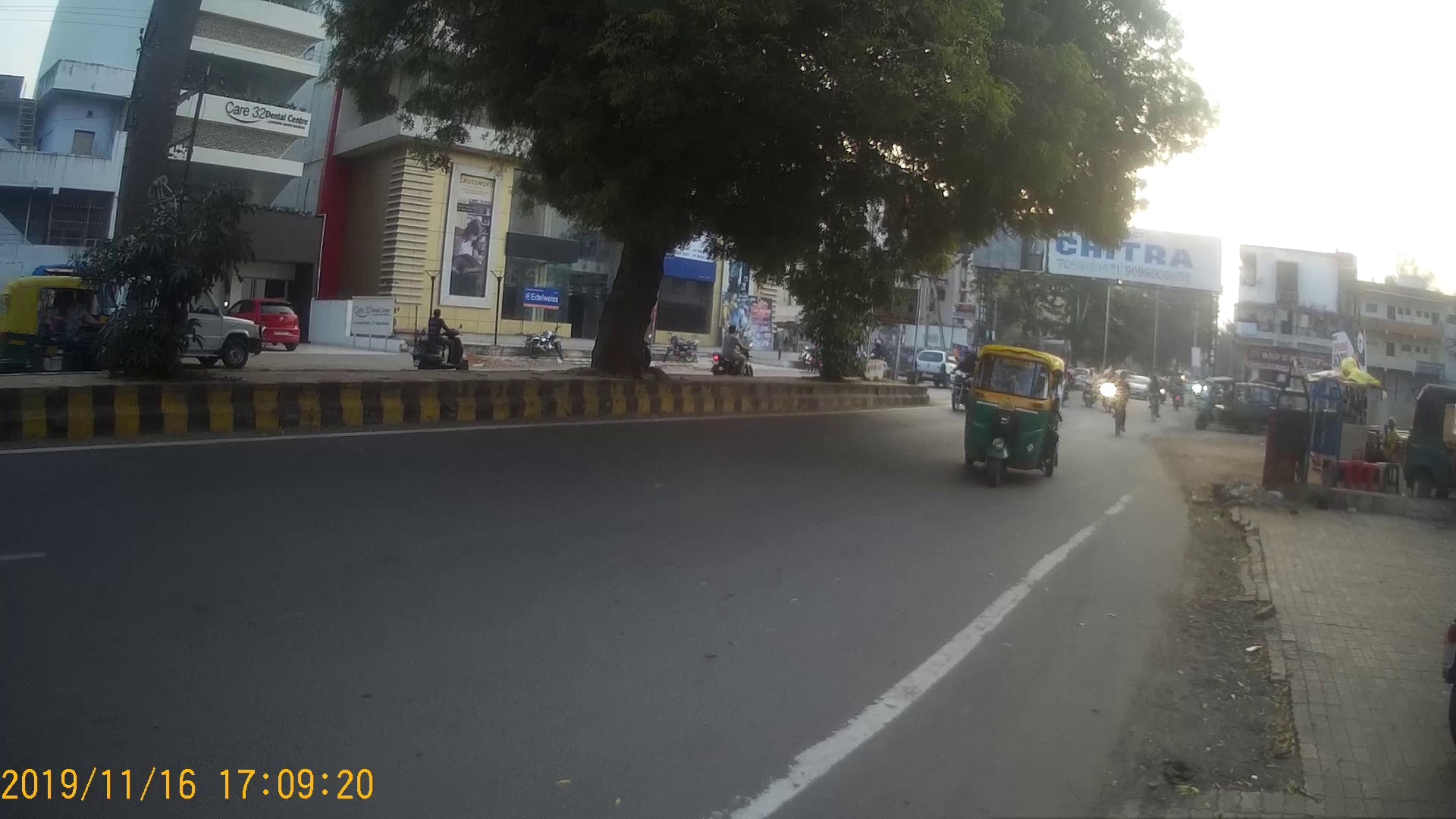}}
    \caption{From top to bottom: sample prediction outputs of MobileNetv2, RoadCrossNet and DilatedRoadCrossNet respectively.}
\end{figure*}

\subsection{Results}

\begin{table}[htp]
  \caption{DL methods: Results on test data}
  \begin{tabular}{c c c c} 
    \toprule
    Method & Precision & Recall & Throughput\\
    \midrule
    MobileNetv2 & 0.90 & 0.60 & 15 fps\\
    RoadCrossNet & 0.90 & 0.72 & 3 fps \\
    DilatedRoadCrossNet & 0.90 & 0.77 & 8 fps \\
  \bottomrule
\end{tabular}
\end{table}
Table 3 shows the precision and recall scores on test set of INDRA achieved by the DL methods as explained in detail above. The table also shows inference throughput obtained on Nvidia Jetson Nano in real-time (deployment details explained in the next section). Sample prediction outputs of the DL methods are shown in Figure 14. We chose DilatedRoadCrossNet for deployment on our working prototype as it exhibited a high precision score, along with decent enough recall score and inference throughput.

\section{Deployment}

\begin{figure*}[htp]
            \centering
            \includegraphics[width=0.8\textwidth]{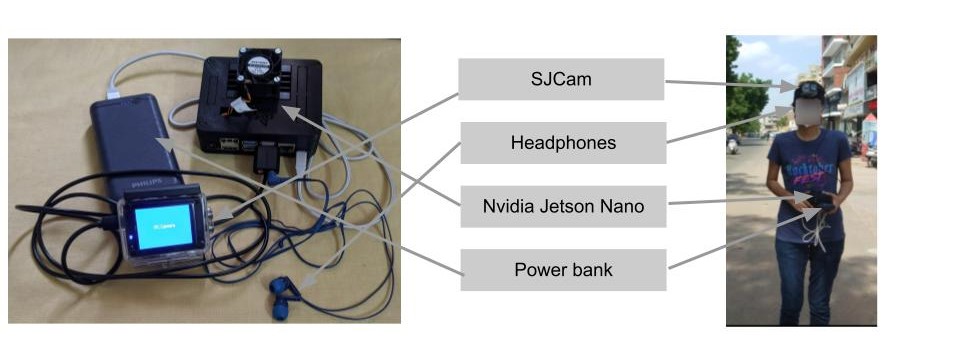}
            \caption{Working prototype of our Road Crossing Assistant}
\end{figure*}

We used Nvidia Jetson Nano and few other supporting components (powerbank, headphones, camera, USB cables \& adapters)  to develop a wearable road-crossing assistant for use in real-time. Low cost is essential for a large-scale impact in developing countries like India. Therefore, we chose to work with a separate camera and mobile computer (Nvidia Jetson Nano) rather than developing a smartphone app. Smartphones with the required deep learning inference hardware are currently significantly more expensive than our discrete combination of components. The workflow of our prototype is as shown in Figure 16.

Firstly we converted our pretrained DilatedRoadCrossNet model into a TensorRT model and deployed it on Jetson Nano. Conversion to TensorRT was essential to optimize the processing power and achieve low latency with high throughput on Jetson Nano. We then wrote a script that is supposed to execute at the start up of Jetson Nano. This script ran the model inference on real-time video stream coming from the camera attached to the Jetson Nano, and also communicated with the user via headphones. Specifically, on startup, our assistant acknowledges the activation through simple audio code and guides the user to look in the direction of oncoming traffic. Next, it runs the deployed pre-trained model on the video stream and communicates with the user when it is safe to cross the road (more precisely, we set a threshold of 0.85, and notify the user to start crossing only when atleast past 5 frames have been predicted as safe). Figure 15 shows the setup of our working prototype. A video demonstrating the use of this assistant in real-time can be found in the supplementary material.
\begin{figure}[htp]
            \centering
            \includegraphics[width=\columnwidth]{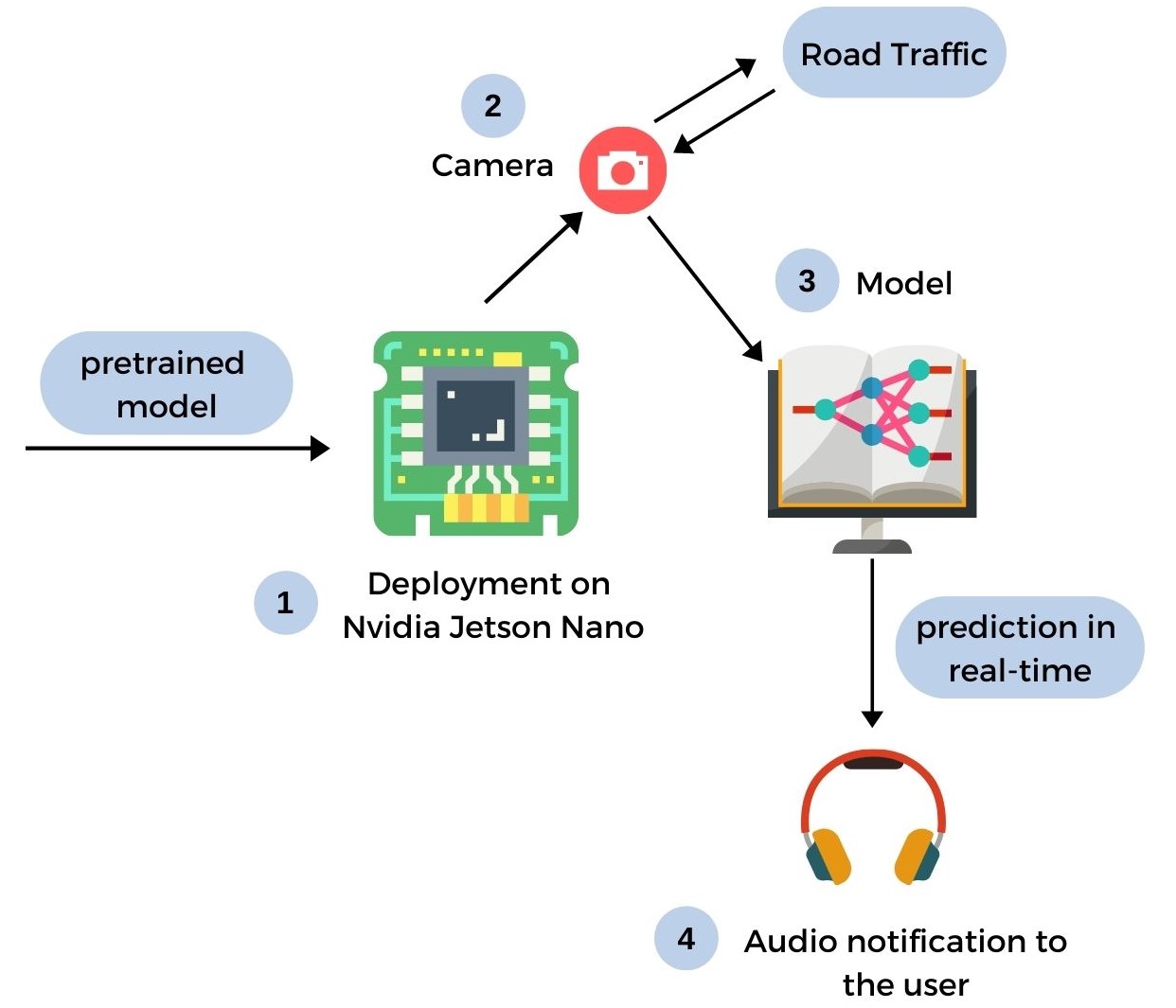}
            \caption{Workflow of our prototype}
\end{figure}

\section{Conclusion and Future Work}

We presented INDRA, the first dataset capturing pedestrian point-of-view traffic patterns of Indian traffic scenarios. We also proposed DilatedRoadCrossNet, which showed promising results for predicting road crossing safety in real-time. Considering that this is the first attempt to train models for predicting road crossing safety from pedestrian point-of-view video stream, our work can serve as a useful baseline for upcoming research in this area. We feel that there are still a lot of ideas worth exploring in order to build a more reliable prediction model. We list some of our future work ideas below:
\begin{enumerate}
    \item Although there was diversity in terms of locations, times, traffic patterns in the dataset, all the videos of this dataset were collected from one city only. We feel that collecting more videos from different cities, different type of roads and over different seasons will add much needed diversity in the dataset. 
    
    \item Multi-frame features: We think that using multi-frame ConvNets will significantly improve the model performance. Recurrent networks appear to be a tempting starting point for most sequence modelling projects, but after a rigorous literature study, we feel that convolutional models can prove out be a better option \cite{bai2018empirical}. Also, since predicting the road crossing safety is an application that requires a lightweight model in order to be deployed for use in real-time, we feel that using convolutional models (capturing spatio-temporal features) over recurrent models (e.g., LSTM, GRU which are comparatively heavier and computationally expensive) is the way to go. We have shortlisted the following approaches which may prove out to be useful for upcoming research in this area:
    
    \begin{enumerate}
        \item Collaborative spatiotemporal feature learning \cite{li2019collaborative}: This approach involves performing 2D convolution along all three orthogonal views of volumetric video data (which learns spatial appearance and temporal motion cues respectively). Also, by sharing the convolution kernels of different views, spatial and temporal features are collaboratively learned and thus benefit from each other; additionally the model doesn't turn out to be very heavy.
        
        \item Using raw frames and optical flow between adjacent frames as two input streams \cite{simonyan2014two, feichtenhofer2017spatiotemporal}: This is a very intuitive approach in video prediction; video can naturally be decomposed into spatial and temporal components. The individual frames carry the spatial information, while the optical flow (between adjacent frames) carries the temporal information. Both the streams are implemented using a convolutional network, and are subsequently combined to predict the class scores.
       
    \end{enumerate}

\end{enumerate}

\begin{acks}

The author(s) are thankful to Govt. of Gujarat’s Startups and Innovation Policy for partial funding of this project, and to AWS Machine Learning Hero Mike Chambers for AWS credits that helped perform deep learning training experiments. We are also thankful to Yagnesh Patil for his help in the deployment phase of this project, and for maintaining the project website. 

\end{acks}